\documentclass[journal]{IEEEtran}
\usepackage{amssymb,amsmath,amsfonts,theorem}
\usepackage{ragged2e}
\usepackage{xspace}
\usepackage{algorithmic}
\usepackage{algorithm}
\usepackage{array}
\usepackage{balance}
\usepackage[section]{placeins}
\usepackage{makecell}
\usepackage[caption=false,font=footnotesize,labelfont=rm, textfont=rm]{subfig}
\usepackage{tabularx,booktabs}
\usepackage{textcomp}
\usepackage{stfloats}
\usepackage{verbatim}
\usepackage{graphicx}
\usepackage{cite}
\usepackage{orcidlink}
\usepackage{float}

\usepackage{longtable}
\graphicspath{{./pdf/}{./jpeg/}{./fig/}{./visio/}}
\DeclareGraphicsExtensions{.pdf,.jpeg,.png,.vsdx}
\def\fig{Fig.}

\usepackage{tikz}
\newcommand*\emptycirc[1][1ex]{\tikz\draw (0,0) circle (#1);} 
\newcommand*\halfcirc[1][1ex]{%
	\begin{tikzpicture}
	\draw[fill] (0,0)-- (90:#1) arc (90:270:#1) -- cycle ;
	\draw (0,0) circle (#1);
	\end{tikzpicture}}
\newcommand*\fullcirc[1][1ex]{\tikz\fill (0,0) circle (#1);} 
\usepackage{array}
\usepackage{makecell}
\usepackage{booktabs}  % 提供 \toprule \midrule \bottomrule
\usepackage{multirow}  % 提供 \multirow 命令
\usepackage{geometry}  % 可根据需要设置页面边距
\usepackage{fontawesome}
\usepackage{pifont}
\usepackage{xcolor} % 用于颜色

\usepackage{orcidlink}

\ifodd 1
\newcommand{\com}[1]{\textbf{\color{red}(COMMENT: #1)}} %comment of the text
\else

\newcommand{\com}[1]{}
\fi

\begin{document}
\justifying
% \title{Generative AI Meets Wireless Sensing: A Paradigm Shift Through Data Generation for Enhanced Performance}
% \title{Generative AI Meets Wireless Sensing: Enhancing Performance through Generative Approaches}
% title{Generative AI Meets Wireless Sensing: Enhancing Performance through Generative Paradigms}
% \title{Generative AI Meets Wireless Sensing: Enhancing Performance through Generative Approaches}
% \title{Generative AI Meets Wireless Sensing: Towards a New Paradigm for Performance Enhancement}
% \title{AIGC Meets Wireless Sensing: Towards Superior Performance via a Generative Paradigm}
% \title{GenAI Meets Wireless Sensing: Unleashing Superior Performance via Generative Paradigms}
% \title{Generative AI Meets Wireless Sensing: Achieving Superior Performance via a Generative Paradigm}
% \title{Generative AI Meets Wireless Sensing: Pipelines, Techniques, and Beyond}
\title{Generative AI Meets Wireless Sensing: Towards Wireless Foundation Model}

\author{Zheng Yang$^{\orcidlink{0000-0003-4048-2684}}$,~\IEEEmembership{Fellow,~IEEE,}
Guoxuan Chi$^{\orcidlink{0000-0002-6941-1612}}$,~\IEEEmembership{Member,~IEEE,}
Chenshu Wu$^{\orcidlink{0000-0002-9700-4627}}$,~\IEEEmembership{Senior Member,~IEEE,} \\ \vspace{-0.5em}
Hanyu Liu$^{\orcidlink{0009-0004-1002-0344}}$,~\IEEEmembership{Student Member,~IEEE,}
Yuchong Gao$^{\orcidlink{0000-0002-4457-4722}}$,~\IEEEmembership{Student Member,~IEEE,}
Yunhao Liu,~\IEEEmembership{Fellow,~IEEE,} \\ \vspace{-0.5em}
Jie Xu$^{\orcidlink{0000-0002-4854-8839}}$,~\IEEEmembership{Fellow,~IEEE,}
Tony Xiao Han,~\IEEEmembership{Senior Member,~IEEE}\\

  \IEEEcompsocitemizethanks{
    \IEEEcompsocthanksitem Zheng Yang, Guoxuan Chi, Hanyu Liu,  Yuchong Gao, and Yunhao Liu are with the Tsinghua University, Beijing, China (e-mail: hmilyyz@gmail.com, chiguoxuan@gmail.com, hanyuliu03@gmail.com, gaoyc01@gmail.com, yunhao@tsinghua.edu.cn).
    \IEEEcompsocthanksitem Chenshu Wu is with the Department of Computer Science, University of Hong Kong, Hong Kong 999077 (e-mail: chenshu@cs.hku.hk).
    \IEEEcompsocthanksitem Jie Xu is with the School of Science and Engineering (SSE), Chinese University of Hong Kong (Shenzhen), Guangdong 518172, China (email: xujie@cuhk.edu.cn).
    \IEEEcompsocthanksitem Tony Xiao Han is with the Huawei Technology Co. Ltd., Guangdong 518129, China (tony.hanxiao@huawei.com).
    \IEEEcompsocthanksitem Zheng Yang is the corresponding author.
  }
}

\maketitle

\begin{abstract}

Generative Artificial Intelligence (GenAI) has made significant advancements in fields such as computer vision (CV) and natural language processing (NLP), demonstrating its capability to synthesize high-fidelity data and improve generalization. Recently, there has been growing interest in integrating GenAI into wireless sensing systems. By leveraging generative techniques such as data augmentation, domain adaptation, and denoising, wireless sensing applications, including device localization, human activity recognition, and environmental monitoring, can be significantly improved.
This survey investigates the convergence of GenAI and wireless sensing from two complementary perspectives. First, we explore how GenAI can be integrated into wireless sensing pipelines, focusing on two modes of integration: as a plugin to augment task-specific models and as a solver to directly address sensing tasks. Second, we analyze the characteristics of mainstream generative models, such as Generative Adversarial Networks (GANs), Variational Autoencoders (VAEs), and diffusion models, and discuss their applicability and unique advantages across various wireless sensing tasks.
We further identify key challenges in applying GenAI to wireless sensing and outline a future direction toward a wireless foundation model—a unified, pre-trained design capable of scalable, adaptable, and efficient signal understanding across diverse sensing tasks.

% Generative Artificial Intelligence has revolutionized fields like computer vision and natural language processing, and its integration into wireless sensing systems has the potential to enhance applications such as object detection, gesture recognition, and environmental monitoring. This survey explores the convergence of GenAI and wireless sensing from two perspectives: (1) integrating GenAI into existing sensing pipelines, either as a plugin to augment traditional models or as a solver to directly address sensing tasks, and (2) analyzing specific GenAI techniques, such as GANs, VAEs, and transformers, and their suitability for tasks like data augmentation and noise reduction in wireless sensing.

% We also discuss challenges, including data scarcity and model generalization, that arise when combining GenAI with wireless sensing. Finally, we propose the concept of a wireless foundation model as the future direction, leveraging GenAI to build scalable, adaptive wireless sensing systems.

\end{abstract}


\section{Introduction}
\label{sec:intro}
\justifying

\begin{figure*}[t]
 \centering
 \includegraphics[width=0.78\linewidth]{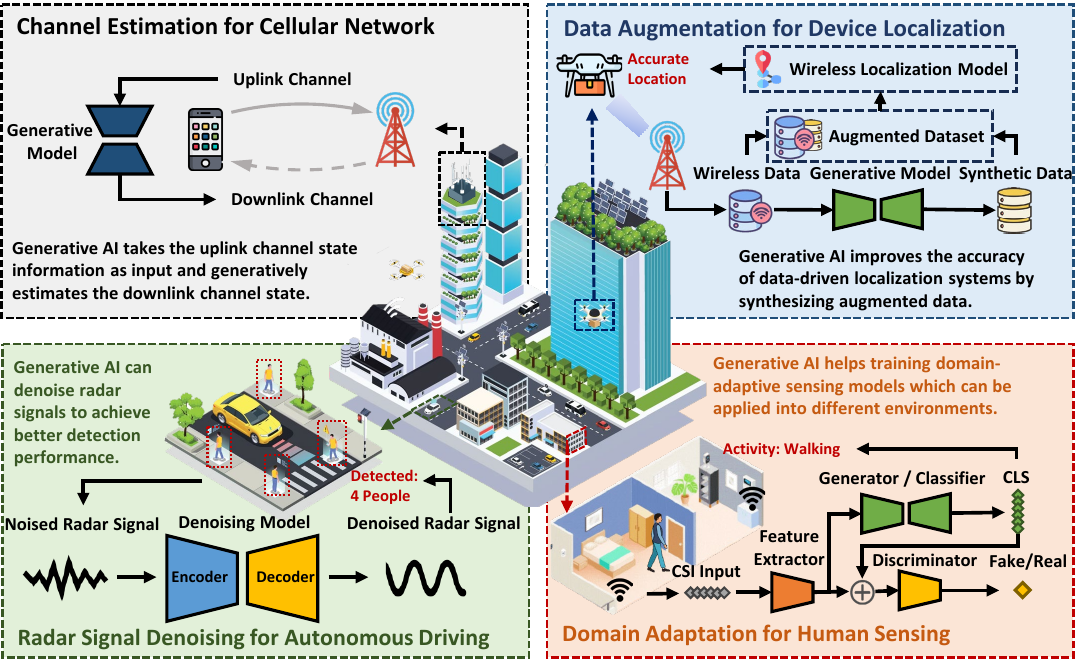}
 
 \caption{\textbf{Typical application scenarios of Generative AI-enabled wireless sensing.} Generative AI enhances the performance of various wireless tasks by leveraging generative methods for channel estimation and signal denoising, as well as serving as a data augmentation or domain adaptation module to improve the accuracy of device localization and human activity perception.}
 \label{fig:motivation-application}
\end{figure*}

\IEEEPARstart{G}{enerative} Artificial Intelligence (GenAI) has emerged as a transformative paradigm in machine learning, offering the ability to learn complex data distributions and generate high-fidelity outputs across modalities. Driven by advancements in deep learning techniques, GenAI has given rise to a new generation of cutting-edge applications, such as Stable Diffusion~\cite{rombach2022high}, Midjourney~\cite{midjourney}, and DALL-E~\cite{ramesh2022hierarchical} for image creation, as well as ChatGPT~\cite{openai2023gpt4} for text generation. These advances have redefined the frontiers of data synthesis, cross-modal reasoning, and zero-shot generalization.

In recent years, GenAI has increasingly been applied to wireless systems. For instance, recent studies have explored the use of GenAI for channel estimation~\cite{zhou2025generative, zheng2024large}, resource allocation~\cite{zheng2024energy}, and beam management~\cite{ghassemi2024multi}, primarily targeting communication-layer tasks. 
Beyond communication, wireless sensing~\cite{li2021deep, liu2019wireless} has emerged as a particularly promising area for GenAI integration. It leverages different types of wireless signals, such as those from Wi-Fi, cellular networks, mmWave radar, and IoT devices to perceive the surrounding environment, enabling tasks such as detection~\cite{zhou2021radio}, estimation~\cite{qian2017widar}, recognition~\cite{zheng2019zero}, and environmental reconstruction~\cite{liu2024wireless}.
It has found increasing adoption in emerging applications such as low-altitude economy, including uncrewed aerial vehicle (UAV) detection~\cite{li2024zero} and tracking~\cite{yu2020conditional}; autonomous driving, where it supports obstacle sensing~\cite{kumar2024advanced} and in-cabin monitoring~\cite{zhang2023vecare}; and smart home environments, which benefit from capabilities like human activity recognition~\cite{wang2021multimodal} and fall detection~\cite{yang2022rethinking}.
A major design trend in this domain is Integrated Sensing and Communication (ISAC)~\cite{liu2022integrated, jin2023integrated},  in which sensing and communication capabilities are jointly optimized within a unified system. This convergence opens new opportunities for applying GenAI models that can jointly exploit signal patterns for both communication and perception tasks.

% Most existing wireless sensing systems still rely on conventional design approaches that are either theoretical model-based or supervised learning-based methods tailored to specific tasks. 
% Theoretical model-based signal processing techniques typically rely on handcrafted features and strong assumptions about propagation environments, such as geometric channel models or simplified multipath effects. While interpretable, these methods often struggle to scale in complex or dynamic scenarios~\cite{}. On the other hand, supervised model designs, particularly deep learning models trained for specific sensing tasks, require large volumes of labeled data and tend to generalize poorly across environments or devices~\cite{}.
% These approaches face several limitations, including data scarcity~\cite{}, lack of model generalization~\cite{}, and the need for task-specific model design~\cite{}. Such challenges hinder the scalability and adaptability of wireless sensing systems, especially in diverse real-world deployments.

% Despite recent advances, most existing wireless sensing systems still rely on  discriminative modeling approaches, which can be broadly categorized into theoretical model-based methods and task-specific supervised learning models. 

Despite recent advances, most wireless sensing systems still rely on conventional discriminative modeling approaches, including model-based signal processing and task-specific supervised learning.
Model-based signal processing techniques typically depend on handcrafted features and strong assumptions about propagation environments, such as geometric channel models or simplified multi-path effects. While interpretable and analytically grounded, these methods often struggle to generalize to complex or dynamic scenarios~\cite{mckown1991ray}.
Supervised learning approaches, particularly deep neural networks trained for specific sensing tasks, require large amounts of labeled data and are prone to overfitting, leading to poor generalization across environments and devices~\cite{dalton2012comparing}.
Overall, these conventional methods face several fundamental limitations, including data scarcity~\cite{konak2023overcoming}, limited generalization~\cite{zou2018robust}, and the need for expert-driven, task-specific model design~\cite{zhang2021wi}. Such challenges significantly hinder the scalability and adaptability of wireless sensing systems in diverse real-world deployments.

To address these issues, generative AI has emerged as a promising paradigm. Recent advances in GenAI, including models such as Generative Adversarial Networks (GANs)~\cite{creswell2018generative}, Variational Autoencoders (VAEs)~\cite{kingma2013auto}, and diffusion models~\cite{sohl2015deep}, offer powerful capabilities in data synthesis, domain adaptation, and robust representation learning across diverse modalities. These strengths make GenAI particularly well-suited to wireless sensing systems, where data is often limited, annotations are costly, and signal conditions vary widely. By generating high-quality synthetic data and modeling complex signal distributions, GenAI can alleviate the dependency on large-scale labeled datasets, enhance cross-domain generalization, and improve robustness to noise and signal distortion.

% Early efforts to integrate GenAI into wireless sensing pipelines have already shown encouraging results across various downstream tasks, such as device localization, human activity recognition, and wireless channel estimation. These empirical studies provide concrete evidence that the combination of GenAI and wireless sensing is not merely a theoretical possibility, but a practical and effective approach for building intelligent, scalable, and robust sensing systems.

\begin{figure*}[t]
\setlength\abovecaptionskip{0pt}
	\centering  %图片全局居中
	\subfloat[Human activity sensing]{
		\includegraphics[width=0.668\linewidth]{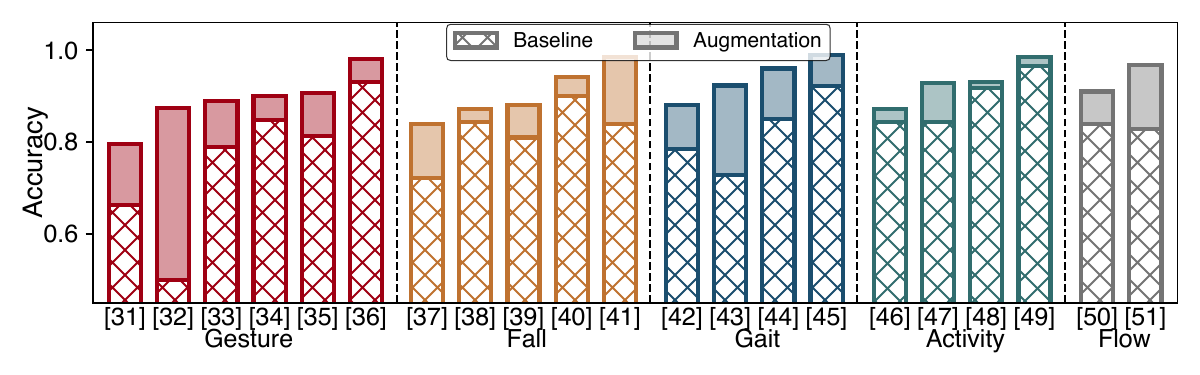}}
	\subfloat[Device localization]{
		\includegraphics[width=0.332\linewidth]{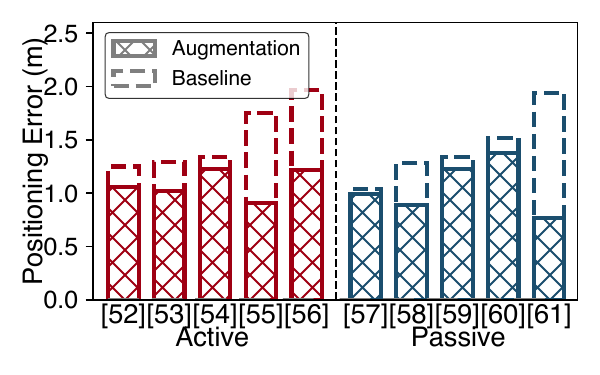}}
        \caption{Performance gains of generative model as a plugin for data augmentation in wireless sensing pipelines.}
        \label{fig:motivation-plugin}
\end{figure*}

\begin{figure*}[t]
\setlength\abovecaptionskip{0pt}
	\centering  %图片全局居中
        \subfloat[Anomaly detection]{
		\includegraphics[width=0.332\linewidth]{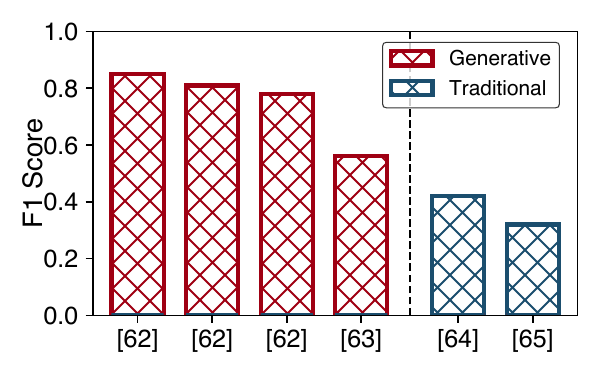}}
	\subfloat[Wireless channel estimation]{
		\includegraphics[width=0.332\linewidth]{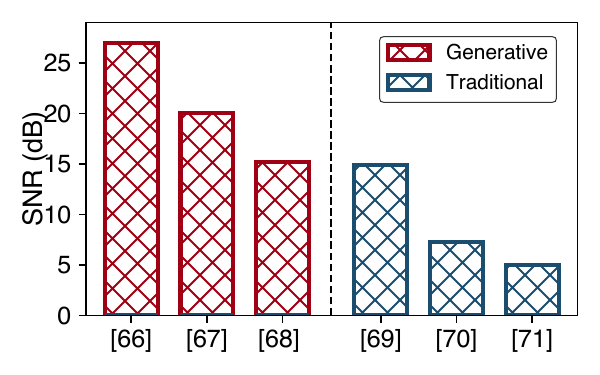}}
        \subfloat[Radio map generation]{
		\includegraphics[width=0.332\linewidth]{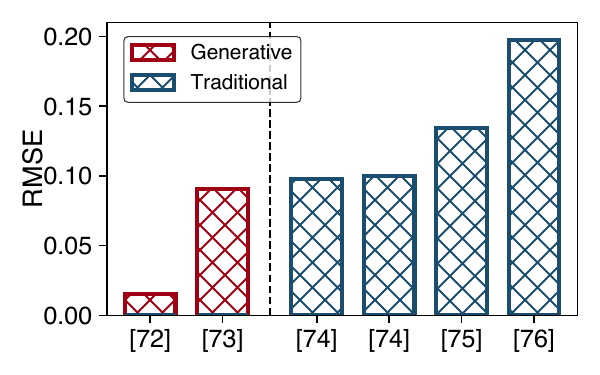}}
        \caption{Performance improvement of generative methods as standalone task solvers compared to traditional methods.}
        \label{fig:motivation-direct}
\end{figure*}

\begin{figure}[t]
     \centering
     \includegraphics[width=0.9\linewidth]{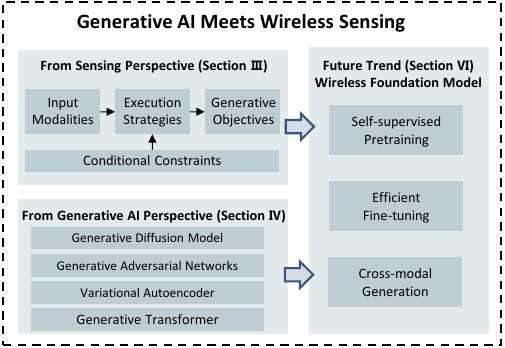}
     \caption{\textbf{Structure of this survey.} This paper reviews the applications of Generative AI in wireless sensing tasks from two distinct perspectives: the wireless sensing pipeline and generative AI techniques. Based on this analysis, this paper further outlines future trends towards wireless foundation models.}
     \label{fig:table-of-contents}
\end{figure}

Early efforts to integrate GenAI into wireless sensing pipelines have shown encouraging results across various downstream tasks. These studies demonstrate that the combination of generative models and wireless sensing is not just a theoretical direction, but a practical and effective approach toward building intelligent and scalable sensing systems. To better understand the current landscape, we categorize existing work into two major paradigms: using GenAI as a plugin to enhance traditional models, and as a solver to directly tackle sensing tasks.

On one hand, \fig\ref{fig:motivation-plugin} illustrates the advantages of using GenAI as a plugin in wireless sensing systems. Specifically, researchers have employed generative models for data augmentation and domain adaptation to enhance existing wireless systems in a wide range of applications such as human activity sensing~\cite{hou2024rfboost, li2021crossgr, wang2020learning, chen2023rf, H6, jiang2020wigan, mao2024wi, H11, 9968957, 9885675, huang2021wiargan, H14, H15, li2024diffgait, H13, moshiri2020using, vishwakarma2021gan, wang2024aigc, huang2023diffar, wang2024generative, yun2023gan} and device localization~\cite{L8, L10, L7, li2018wavelet, L9, L5, L4, L3, L1, 9374989}. 
These improvements have led to substantial performance gains across a range of wireless sensing tasks, as demonstrated in recent studies on gesture recognition~\cite{hou2024rfboost, li2021crossgr, wang2020learning, chen2023rf, H6, jiang2020wigan}, fall detection~\cite{mao2024wi, H11, 9968957, 9885675, huang2021wiargan}, gait analysis~\cite{H14,H15,li2024diffgait,H13}, activity recognition~\cite{moshiri2020using, vishwakarma2021gan, wang2024aigc, huang2023diffar},  and crowd detection~\cite{wang2024generative,yun2023gan}, as reported in the original studies.

On the other hand, \fig\ref{fig:motivation-direct} demonstrates the effectiveness of using GenAI models as a standalone task solver. Compared to traditional non-generative approaches, GenAI has shown superior performance in representative tasks such as signal anomaly detection~\cite{10017520,schlegl2019f,liu2008isolation,wang2004anomaly}, channel estimation~\cite{chi2024rf,zhao2023nerf2,liu2021fire,vasisht2016eliminating,bakshi2019fast,kaltenberger2008performance}, and radio map generation~\cite{RME-GAN,goodfellow2020generative,teganya2021deep,krumm2003minimizing,lee2012voronoi}. These results further highlight the powerful potential of GenAI in advancing a variety of wireless tasks. These performance improvements are also drawn directly from the experimental results reported in the cited works.

In view of the growing trends, this paper provides a comprehensive survey on the integration of GenAI into wireless sensing, as illustrated in \fig\ref{fig:table-of-contents}. Section~\ref{sec:related} reviews the current state of research and identifies gaps that demonstrate the promise of GenAI in wireless sensing. Section~\ref{sec:sensing} presents two complementary perspectives on how GenAI can be integrated into wireless sensing pipelines: as a plugin to enhance traditional models and as a solver to directly address sensing tasks. Section~\ref{sec:generative} discusses specific GenAI techniques, such as GANs, VAEs, and diffusion models, evaluating their strengths and limitations in practical wireless sensing applications. Section~\ref{sec:issue} analyzes key challenges in integrating GenAI, including data scarcity, model generalization, and scalability, which remain open problems in the field. Finally, Section~\ref{sec:future} proposes a future vision of a wireless foundation model that unifies these GenAI techniques to create scalable and adaptable systems, and the paper concludes in Section~\ref{sec:conclusion}.

\section{Related Work}
\label{sec:related}
% \justifying

% \begin{figure*}[b]
%      \centering
%      \includegraphics[width=1\linewidth]{fig/5-generative.pdf}
%      \caption{From traditional machine learning to generative AI: key milestones and four representative generative model architectures.}
%      \label{fig:generative}
%  \end{figure*}

\renewcommand\arraystretch{1.5}
\begin{table*}[t]

\caption{Comparison between existing surveys on generative model for wireless applications}
\begin{tabular}{|c|ccc|cccc|p{7.4cm}|}
\hline
\multirow{2}{*}{Ref.}    & \multicolumn{3}{c|}{Modalities} & \multicolumn{4}{c|}{Generative Methods} & \multicolumn{1}{c|}{\multirow{2}{*}{Scope \& Key Techniques}} \\ \cline{2-8}
                         & Cellular   & Radar   & Wi-Fi   & GAN  & VAE  & Diffusion  & AR & \multicolumn{1}{c|}{}                                  \\ \hline

 \cite{10490142} &\textcolor[RGB]{77,168,67}{\ding{51}}              &\textcolor[RGB]{197,5,56}{\ding{55}}         &\textcolor[RGB]{77,168,67}{\ding{51}}         &\textcolor[RGB]{77,168,67}{\ding{51}}       &\textcolor[RGB]{77,168,67}{\ding{51}}       & \textcolor[RGB]{77,168,67}{\ding{51}}            &\textcolor[RGB]{77,168,67}{\ding{51}}             & Communication, GenAI,  Channel Estimation \& Equalization                                        \\\hline
\cite{fahimekhoramnejad} & \textcolor[RGB]{77,168,67}{\ding{51}}            & \textcolor[RGB]{77,168,67}{\ding{51}}         & \textcolor[RGB]{197,5,56}{\ding{55}}          &\textcolor[RGB]{77,168,67}{\ding{51}}       &\textcolor[RGB]{197,5,56}{\ding{55}}      & \textcolor[RGB]{197,5,56}{\ding{55}}            & \textcolor[RGB]{77,168,67}{\ding{51}}             &Communication, GenAI,  xG Optimization \& Load Balancing                                                        \\ \hline
\cite{10614204} &  \textcolor[RGB]{77,168,67}{\ding{51}}            & \textcolor[RGB]{197,5,56}{\ding{55}}           & \textcolor[RGB]{197,5,56}{\ding{55}}           &  \textcolor[RGB]{77,168,67}{\ding{51}}      &  \textcolor[RGB]{77,168,67}{\ding{51}}      &  \textcolor[RGB]{77,168,67}{\ding{51}}            & \textcolor[RGB]{77,168,67}{\ding{51}}              & Communication, GenAI, Resource Allocation      \\ \hline
\cite{9726814} &  \textcolor[RGB]{197,5,56}{\ding{55}}    &  \textcolor[RGB]{197,5,56}{\ding{55}}       &   \textcolor[RGB]{197,5,56}{\ding{55}}      &  \textcolor[RGB]{77,168,67}{\ding{51}}    &  \textcolor[RGB]{77,168,67}{\ding{51}}    &    \textcolor[RGB]{197,5,56}{\ding{55}}        &         \textcolor[RGB]{197,5,56}{\ding{55}}    & IIoT, GenAI, Anomaly Detection, Platform Monitoring                           \\ \hline
\cite{10384630} &\textcolor[RGB]{77,168,67}{\ding{51}}              & \textcolor[RGB]{197,5,56}{\ding{55}}         & \textcolor[RGB]{77,168,67}{\ding{51}}          &\textcolor[RGB]{197,5,56}{\ding{55}}       &\textcolor[RGB]{197,5,56}{\ding{55}}       & \textcolor[RGB]{197,5,56}{\ding{55}}            & \textcolor[RGB]{77,168,67}{\ding{51}}              &Communication, DL, 6G Autonomous \& Self-Evolving Networks                                        \\ \hline
 \cite{10623395} & \textcolor[RGB]{197,5,56}{\ding{55}}              & \textcolor[RGB]{197,5,56}{\ding{55}}           &\textcolor[RGB]{197,5,56}{\ding{55}}            & \textcolor[RGB]{77,168,67}{\ding{51}}        & \textcolor[RGB]{77,168,67}{\ding{51}}      & \textcolor[RGB]{77,168,67}{\ding{51}}             & \textcolor[RGB]{197,5,56}{\ding{55}}            &  Communication, GenAI, Security \& Authentication                                                      \\ \hline
\cite{Lexia} &\textcolor[RGB]{77,168,67}{\ding{51}}              &\textcolor[RGB]{197,5,56}{\ding{55}}         &\textcolor[RGB]{197,5,56}{\ding{55}}        &\textcolor[RGB]{197,5,56}{\ding{55}}       & \textcolor[RGB]{197,5,56}{\ding{55}}              & \textcolor[RGB]{77,168,67}{\ding{51}}& \textcolor[RGB]{77,168,67}{\ding{51}}             &Communication, GenAI,  Multimodal Content Provisioning                                  \\ \hline

\cite{10.1145/3436729} &\textcolor[RGB]{197,5,56}{\ding{55}}            &\textcolor[RGB]{77,168,67}{\ding{51}}          &\textcolor[RGB]{77,168,67}{\ding{51}}          &\textcolor[RGB]{77,168,67}{\ding{51}}       &\textcolor[RGB]{197,5,56}{\ding{55}}       &\textcolor[RGB]{197,5,56}{\ding{55}}             & \textcolor[RGB]{77,168,67}{\ding{51}}               &Sensing, DL, Feature Extraction, Localization          \\ \hline
\cite{8794643} &\textcolor[RGB]{197,5,56}{\ding{55}}            &\textcolor[RGB]{77,168,67}{\ding{51}}         &\textcolor[RGB]{77,168,67}{\ding{51}}         &\textcolor[RGB]{197,5,56}{\ding{55}}       &\textcolor[RGB]{197,5,56}{\ding{55}}       &\textcolor[RGB]{197,5,56}{\ding{55}}             & \textcolor[RGB]{197,5,56}{\ding{55}}             & Sensing, ML, Activity Recognition                                          \\ \hline
\cite{8284052} &\textcolor[RGB]{197,5,56}{\ding{55}}             &\textcolor[RGB]{77,168,67}{\ding{51}}          &\textcolor[RGB]{77,168,67}{\ding{51}}          &\textcolor[RGB]{197,5,56}{\ding{55}}       &\textcolor[RGB]{197,5,56}{\ding{55}}       &\textcolor[RGB]{197,5,56}{\ding{55}}             &\textcolor[RGB]{197,5,56}{\ding{55}}              &Sensing, Device-Free, Activity Recognition, Localization                                                \\  \hline
% \cite{9076119} &\textcolor[RGB]{197,5,56}{\ding{55}}             &\textcolor[RGB]{77,168,67}{\ding{51}}          &\textcolor[RGB]{77,168,67}{\ding{51}}          &\textcolor[RGB]{77,168,67}{\ding{51}}       &\textcolor[RGB]{197,5,56}{\ding{55}}       & \textcolor[RGB]{197,5,56}{\ding{55}}            &\textcolor[RGB]{197,5,56}{\ding{55}}              & \textbf{ Sensing Tasks like Location, Gestures, and Vital Signs.}                                                      \\
% \textbf{Ours}            &    \textcolor[RGB]{77,168,67}{\ding{51}}        &   \textcolor[RGB]{77,168,67}{\ding{51}}      &  \textcolor[RGB]{77,168,67}{\ding{51}}       &  \textcolor[RGB]{77,168,67}{\ding{51}}    &  \textcolor[RGB]{77,168,67}{\ding{51}}    &   \textcolor[RGB]{77,168,67}{\ding{51}}         &   \textcolor[RGB]{77,168,67}{\ding{51}}          & \textbf{Sensing:} comprehensive survey on the application of various generative AI models across diverse wireless sensing tasks      
\cite{feng2022survey} &\textcolor[RGB]{197,5,56}{\ding{55}}             &\textcolor[RGB]{197,5,56}{\ding{55}}          &\textcolor[RGB]{77,168,67}{\ding{51}}          &\textcolor[RGB]{77,168,67}{\ding{51}}       &\textcolor[RGB]{197,5,56}{\ding{55}}       &\textcolor[RGB]{197,5,56}{\ding{55}}             &\textcolor[RGB]{197,5,56}{\ding{55}}              & Sensing, DL, Localization                                                    \\  \hline

\cite{yousefi2017survey} &\textcolor[RGB]{197,5,56}{\ding{55}}             &\textcolor[RGB]{197,5,56}{\ding{55}}          &\textcolor[RGB]{77,168,67}{\ding{51}}          &\textcolor[RGB]{197,5,56}{\ding{55}}        &\textcolor[RGB]{197,5,56}{\ding{55}}       &\textcolor[RGB]{197,5,56}{\ding{55}}             &\textcolor[RGB]{197,5,56}{\ding{55}}              & Sensing, DL, Activity Recognition, Feature Extraction                                                    \\  \hline

\cite{feng2025survey} &\textcolor[RGB]{197,5,56}{\ding{55}}             &\textcolor[RGB]{197,5,56}{\ding{55}}          &\textcolor[RGB]{77,168,67}{\ding{51}}          &\textcolor[RGB]{77,168,67}{\ding{51}}        &\textcolor[RGB]{77,168,67}{\ding{51}}       &\textcolor[RGB]{197,5,56}{\ding{55}}             &\textcolor[RGB]{197,5,56}{\ding{55}}              & Sensing, DL, Indoor Localization                                                  \\  \hline

\cite{wang2019survey} &\textcolor[RGB]{197,5,56}{\ding{55}}             &\textcolor[RGB]{197,5,56}{\ding{55}}          &\textcolor[RGB]{77,168,67}{\ding{51}}          &\textcolor[RGB]{197,5,56}{\ding{55}}        &\textcolor[RGB]{197,5,56}{\ding{55}}       &\textcolor[RGB]{197,5,56}{\ding{55}}             &\textcolor[RGB]{197,5,56}{\ding{55}}              & Sensing, ML, Activity Recognition                                             \\  \hline

\cite{aljumaily2016survey} &\textcolor[RGB]{197,5,56}{\ding{55}}             &\textcolor[RGB]{197,5,56}{\ding{55}}     &\textcolor[RGB]{77,168,67}{\ding{51}}              &\textcolor[RGB]{197,5,56}{\ding{55}}        &\textcolor[RGB]{197,5,56}{\ding{55}}       &\textcolor[RGB]{197,5,56}{\ding{55}}             &\textcolor[RGB]{197,5,56}{\ding{55}}             & Sensing, ML, Activity Recognition                                       \\  \hline

\cite{wei2025survey} &\textcolor[RGB]{197,5,56}{\ding{55}}             &\textcolor[RGB]{197,5,56}{\ding{55}}     &\textcolor[RGB]{77,168,67}{\ding{51}}              &\textcolor[RGB]{77,168,67}{\ding{51}}         &\textcolor[RGB]{197,5,56}{\ding{55}}       &\textcolor[RGB]{197,5,56}{\ding{55}}             &\textcolor[RGB]{197,5,56}{\ding{55}}             & Sensing, GenAI, Human Identification                                             \\  \hline

\cite{wang2024survey} &\textcolor[RGB]{197,5,56}{\ding{55}}             &\textcolor[RGB]{197,5,56}{\ding{55}}     &\textcolor[RGB]{77,168,67}{\ding{51}}              &\textcolor[RGB]{197,5,56}{\ding{55}}         &\textcolor[RGB]{197,5,56}{\ding{55}}       &\textcolor[RGB]{197,5,56}{\ding{55}}             &\textcolor[RGB]{197,5,56}{\ding{55}}             & Sensing, DL, Localization                                                   \\  \hline

\cite{ahmad2024wifi} &\textcolor[RGB]{197,5,56}{\ding{55}}             &\textcolor[RGB]{197,5,56}{\ding{55}}     &\textcolor[RGB]{77,168,67}{\ding{51}}              &\textcolor[RGB]{197,5,56}{\ding{55}}         &\textcolor[RGB]{197,5,56}{\ding{55}}       &\textcolor[RGB]{197,5,56}{\ding{55}}             &\textcolor[RGB]{197,5,56}{\ding{55}}             & Sensing, DL, Activity Recognition, Localization                                     \\  \hline

\cite{ma2019wifi} &\textcolor[RGB]{197,5,56}{\ding{55}}             &\textcolor[RGB]{197,5,56}{\ding{55}}     &\textcolor[RGB]{77,168,67}{\ding{51}}              &\textcolor[RGB]{77,168,67}{\ding{51}}         &\textcolor[RGB]{197,5,56}{\ding{55}}       &\textcolor[RGB]{197,5,56}{\ding{55}}             &\textcolor[RGB]{197,5,56}{\ding{55}}             & Sensing, DL, Activity Recognition, Human Identification                                      \\  \hline

\textbf{Ours}            &    \textcolor[RGB]{77,168,67}{\ding{51}}        &   \textcolor[RGB]{77,168,67}{\ding{51}}      &  \textcolor[RGB]{77,168,67}{\ding{51}}       &  \textcolor[RGB]{77,168,67}{\ding{51}}    &  \textcolor[RGB]{77,168,67}{\ding{51}}    &   \textcolor[RGB]{77,168,67}{\ding{51}}         &   \textcolor[RGB]{77,168,67}{\ding{51}}          & \begin{tabular}[c]{@{}l@{}}Sensing, comprehensive survey on the application of various \\ generative AI models across diverse wireless sensing tasks\end{tabular} \\ \hline
\end{tabular}
\label{tab:related}
\end{table*}

% \usepackage{multirow}
% \usepackage{multirow}

% Table~\ref{tab:related} highlights the differences between our survey and existing works in the field. Most of the current surveys on Generative AI for wireless systems primarily focus on communication scenarios, with a strong emphasis on using generative models for network scheduling and communication system optimization. Some works have explored the use of diffusion models for physical layer signal processing, but these studies still concentrate on improving the throughput and security of wireless communication systems.
% In contrast, existing surveys in the wireless sensing domain often lacks an in-depth investigation into generative models. While some recent surveys briefly mention classical generative models, such as Generative Adversarial Networks (GANs), there is a significant gap when it comes to the latest advancements in generative methods. Additionally, existing surveys often fail to comprehensively cover various wireless signal modalities, further limiting their applicability.

% Our work is the first to provide a thorough investigation of mainstream wireless signal modalities and generative methods in the context of wireless sensing applications. This work fills the gap by offering a detailed overview of how generative AI can be applied to wireless sensing tasks, marking a significant step forward in this emerging field.

Table~\ref{tab:related} summarizes the key distinctions between our survey and existing surveys across several dimensions, including application scope, generative methods, and signal modalities. Most existing surveys on GenAI in wireless contexts predominantly target communication scenarios, emphasizing applications such as network scheduling and system-level optimization~\cite{10490142,fahimekhoramnejad,10384630}. Although a few recent studies investigate diffusion models for physical-layer signal processing~\cite{10614204,10623395}, their focus remains centered on enhancing throughput and security rather than addressing sensing-specific objectives.

In contrast, surveys that focus on wireless sensing often lack a thorough and dedicated analysis of generative models~\cite{10.1145/3436729,8794643,8284052}, even though interest in this area has been steadily increasing. While some recent reviews briefly discuss classical generative models, such as GANs~\cite{10.1145/3436729}, very few offer a systematic overview of more recent developments. These developments include diffusion-based generative architectures and transformer-based sequence models. In addition, the variety of wireless signal inputs, such as Channel State Information (CSI), and radar spectrograms, is typically underexplored in the literature, which reduces the applicability of these surveys to real-world sensing scenarios.

To the best of our knowledge, this survey is the first to offer a comprehensive and structured investigation of generative AI techniques within the context of wireless sensing. It distinguishes itself by jointly covering the diversity of signal modalities and the breadth of generative model families. In addition, existing works lack a clear taxonomy on how generative models can be integrated into sensing pipelines, a gap this paper addresses through the plugin/solver framework. By bridging these two perspectives, our work fills a critical gap in the current literature and provides a foundational reference for future research in this emerging field.

\begin{figure*}[t]
\setlength\abovecaptionskip{5pt}
	\centering  %图片全局居中
	\subfloat[Traditional wireless sensing pipeline]{
        \label{fig:pipeline-traditional}
		\includegraphics[width=0.332\linewidth]{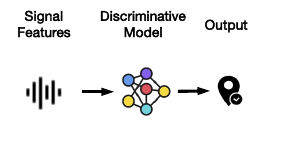}}
	\subfloat[GenAI as plugin]{
        \label{fig:pipeline-plugin}
		\includegraphics[width=0.332\linewidth]{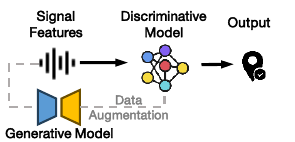}}
        \subfloat[GenAI as standalone solver]{
        \label{fig:pipeline-sovler}
		\includegraphics[width=0.332\linewidth]{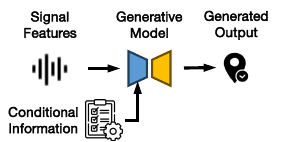}}
        \caption{\textbf{Comparison of wireless sensing pipelines.} (a) The traditional pipeline employs task-specific discriminative models without generative assistance. (b) In plugin-based pipelines, GenAI modules are used to enhance data diversity and domain generalization. (c) In solver-based pipelines, GenAI models directly perform sensing tasks by generating outputs from input signals and conditional priors.}
        \label{fig:pipeline}
\end{figure*}

\section{Wireless Sensing Pipelines with Generative AI}
\label{sec:sensing}

\begin{table*}[p]
\centering
\caption{Representative Wireless Sensing Applications Leveraging Generative AI}
\begin{flushleft}
\textbf{Legend:} 
\fullcirc~= GenAI used as \textbf{task solver};\quad 
\emptycirc~= GenAI used as \textbf{plugin or auxiliary module};\quad 
\halfcirc~= GenAI used in \textbf{both modes}.
\end{flushleft}
\label{tab:example}

\small 
% 使用 p{固定宽度} 并在需要处加上 >{\raggedright} 或 >{\centering} 等控制对齐方式
\renewcommand\arraystretch{1.5}
\begin{tabular}{
  >{\centering\arraybackslash}m{2.0cm}|  % 1. System Name
  >{\centering\arraybackslash}m{2.3cm}|  
  %\raggedright
  % 2. Input (Description)
  >{\centering\arraybackslash}m{0.9cm}|    % 3. Input (Raw Signal)
  >{\centering\arraybackslash}m{2.3cm}|  % 4. Downstream Tasks (Description)
  >{\centering\arraybackslash}m{0.9cm}|   % 5. Downstream Tasks (Task types)
  >{\centering\arraybackslash}m{2.3cm}|  % 6. Generation's Target (Description)
  >{\centering\arraybackslash}m{0.9cm}|    % 7. Generation's Target (Multimodal)
  >{\centering\arraybackslash}m{1.4cm}    % 8. Conditional Generation
}
\hline
\multirow{2}{*}[-0.6em]{\centering \textbf{System Name}} 
& \multicolumn{2}{c|}{\textbf{Input}} 
& \multicolumn{2}{c|}{\textbf{Downstream Tasks}}
& \multicolumn{2}{c|}{\textbf{Generation's Target}}
& \multirow{2}{*}[-0.6em]{\centering \textbf{Condition}} \\
\cline{2-3} \cline{4-5} \cline{6-7}
& \centering \textbf{Description} 
& \centering \textbf{Raw Signal} 
& \centering \textbf{Description} 
& \centering \textbf{Task types} 
& \centering \textbf{Description} 
& \centering \textbf{Multi modal} 
& \\
\hline
RF-Diffusion~\cite{chi2024rf}            
& Wi-Fi CSI, Radar
& \textcolor[RGB]{77,168,67}{\ding{51}}
& Data Completion, Channel Prediction
& \halfcirc
& Real Signal Generation, Network Optimization
& \textcolor[RGB]{197,5,56}{\ding{55}}
& \textcolor[RGB]{77,168,67}{\ding{51}} \\
\hline
SynMotion~\cite{zhang2022synthesized}                
& Visual Data, Human Body Mesh
& \textcolor[RGB]{77,168,67}{\ding{51}}
& Data Completion, Anomaly Detection
& \halfcirc
& Real Signal Generation, Environmental Reconstruction
& \textcolor[RGB]{197,5,56}{\ding{55}}
& \textcolor[RGB]{77,168,67}{\ding{51}} \\
\hline
NeRF\textasciicircum{}2~\cite{zhao2023nerf2} 
& Signal Amplitude/Phase, Environmental Information
& \textcolor[RGB]{197,5,56}{\ding{55}}
& Data Augmentation, Feature Optimization
& \halfcirc
& Environmental Reconstruction
& \textcolor[RGB]{197,5,56}{\ding{55}}
& \textcolor[RGB]{77,168,67}{\ding{51}} \\
\hline
FallDar~\cite{yang2022rethinking}                 
& WiFi-CSI
& \textcolor[RGB]{77,168,67}{\ding{51}}
& Data Augmentation
& \emptycirc
& Feature Generation
& \textcolor[RGB]{197,5,56}{\ding{55}}
& \textcolor[RGB]{77,168,67}{\ding{51}} \\
\hline
Eren Balevi et al.~\cite{balevi2021wideband}                    
& CSI
& \textcolor[RGB]{77,168,67}{\ding{51}}
& Channel Prediction
& \fullcirc
& Real signal generation
& \textcolor[RGB]{197,5,56}{\ding{55}}
& \textcolor[RGB]{77,168,67}{\ding{51}} \\
\hline
Michael Baur et al.~\cite{baur2022variational}                
& CSI
& \textcolor[RGB]{77,168,67}{\ding{51}}
& Channel Prediction
& \fullcirc
& Feature Generation
& \textcolor[RGB]{197,5,56}{\ding{55}}
& \textcolor[RGB]{77,168,67}{\ding{51}} \\
\hline
RF Genesis~\cite{chen2023rf}              
& Visual Data, Graphical Data 
& \textcolor[RGB]{197,5,56}{\ding{55}}
& Data Augmentation, Domain Generalization
& \emptycirc
& Cross-Modal Data Generation
& \textcolor[RGB]{77,168,67}{\ding{51}}
& \textcolor[RGB]{77,168,67}{\ding{51}} \\
\hline
CDDM~\cite{wu2023cddm}                    
& CSI
& \textcolor[RGB]{77,168,67}{\ding{51}}
& Feature Optimization
& \fullcirc
& Feature Generation
& \textcolor[RGB]{197,5,56}{\ding{55}}
& \textcolor[RGB]{77,168,67}{\ding{51}} \\
\hline
DiffGait~\cite{li2024diffgait}                
& Radar, Doppler Shift
& \textcolor[RGB]{77,168,67}{\ding{51}}
& Data Augmentation
& \emptycirc
& Feature Generation
& \textcolor[RGB]{197,5,56}{\ding{55}}
& \textcolor[RGB]{77,168,67}{\ding{51}} \\
\hline
DiffAR~\cite{huang2023diffar}                  
& Wi-Fi CSI, Time-Frequency Characteristics
& \textcolor[RGB]{77,168,67}{\ding{51}}
& Data Completion
& \emptycirc
& Real Signal Generation
& \textcolor[RGB]{197,5,56}{\ding{55}}
& \textcolor[RGB]{77,168,67}{\ding{51}} \\
\hline
RFID-ACCLDM~\cite{wang2024aigc}             
& RFID
& \textcolor[RGB]{77,168,67}{\ding{51}}
& Data Completion, Data Augmentation
& \emptycirc
& Feature Generation, Real Signal Generation
& \textcolor[RGB]{197,5,56}{\ding{55}}
& \textcolor[RGB]{77,168,67}{\ding{51}} \\
\hline
mmDiff~\cite{fan2025diffusion}                  
& Radar Point Cloud
& \textcolor[RGB]{77,168,67}{\ding{51}}
& Feature Optimization
& \fullcirc
& Feature Generation
& \textcolor[RGB]{197,5,56}{\ding{55}}
& \textcolor[RGB]{77,168,67}{\ding{51}} \\
\hline
Wi-Cro~\cite{mao2024wi}                  
& Wi-Fi CSI
& \textcolor[RGB]{77,168,67}{\ding{51}}
& Domain Generalization, Data Augmentation
& \emptycirc
& Real Signal Generation
& \textcolor[RGB]{197,5,56}{\ding{55}}
& \textcolor[RGB]{77,168,67}{\ding{51}} \\
\hline
S. Vishwakarma et al.~\cite{vishwakarma2021gan}                    
& Radar, Noise Features
& \textcolor[RGB]{77,168,67}{\ding{51}}
& Data Completion
& \emptycirc
& Real Signal Generation
& \textcolor[RGB]{197,5,56}{\ding{55}}
& \textcolor[RGB]{77,168,67}{\ding{51}} \\
\hline
J. H. Seong et al.~\cite{seong2019selective}                    
& AP Coordinates, Environmental Information, Wi-Fi RSSI
& \textcolor[RGB]{77,168,67}{\ding{51}}
& Data Completion, Feature Optimization, Image Generation
& \halfcirc
& Feature Generation, Environmental Reconstruction
& \textcolor[RGB]{197,5,56}{\ding{55}}
& \textcolor[RGB]{197,5,56}{\ding{55}} \\
\hline
P. F. Moshiri et al.~\cite{moshiri2020using}                    
& Wi-Fi CSI
& \textcolor[RGB]{77,168,67}{\ding{51}}
& Data Augmentation
& \emptycirc
& Feature Generation
& \textcolor[RGB]{197,5,56}{\ding{55}}
& \textcolor[RGB]{77,168,67}{\ding{51}} \\
\bottomrule
\end{tabular}
\end{table*}

\begin{table*}[htbp]
\centering
\addtocounter{table}{-1}
\caption{Representative Wireless Sensing Applications Leveraging Generative AI (Continued)}
\label{tab:example-continue}

\small 
% 使用 p{固定宽度} 并在需要处加上 >{\raggedright} 或 >{\centering} 等控制对齐方式
\renewcommand\arraystretch{1.3}
\begin{tabular}{
  >{\centering\arraybackslash}m{2.0cm}|  % 1. System Name
  >{\centering\arraybackslash}m{2.3cm}|  
  %\raggedright
  % 2. Input (Description)
  >{\centering\arraybackslash}m{0.9cm}|    % 3. Input (Raw Signal)
  >{\centering\arraybackslash}m{2.3cm}|  % 4. Downstream Tasks (Description)
  >{\centering\arraybackslash}m{0.9cm}|   % 5. Downstream Tasks (Task types)
  >{\centering\arraybackslash}m{2.3cm}|  % 6. Generation's Target (Description)
  >{\centering\arraybackslash}m{0.9cm}|    % 7. Generation's Target (Multimodal)
  >{\centering\arraybackslash}m{1.4cm}    % 8. Conditional Generation
}
\hline
\multirow{2}{*}[-0.6em]{\centering \textbf{System Name}} 
& \multicolumn{2}{c|}{\textbf{Input}} 
& \multicolumn{2}{c|}{\textbf{Downstream Tasks}}
& \multicolumn{2}{c|}{\textbf{Generation's Target}}
& \multirow{2}{*}[-0.6em]{\centering \textbf{Condition}} \\
\cline{2-3} \cline{4-5} \cline{6-7}
& \centering \textbf{Description} 
& \centering \textbf{Raw Signal} 
& \centering \textbf{Description} 
& \centering \textbf{Task types} 
& \centering \textbf{Description} 
& \centering \textbf{Multi modal} 
& \\
\hline
C. Lin et al.~\cite{lin2024rf}                    
& Wi-Fi CSI
& \textcolor[RGB]{77,168,67}{\ding{51}}
& Data Augmentation
& \emptycirc
& Network Optimization, Real Signal Generation
& \textcolor[RGB]{197,5,56}{\ding{55}}
& \textcolor[RGB]{77,168,67}{\ding{51}} \\
\hline
DANGR~\cite{han2020deep}                   
& Wi-Fi CSI
& \textcolor[RGB]{77,168,67}{\ding{51}}
& Domain Generalization, Data Augmentation
& \emptycirc
& Real Signal Generation
& \textcolor[RGB]{197,5,56}{\ding{55}}
& \textcolor[RGB]{77,168,67}{\ding{51}} \\
\hline
WiADG~\cite{zou2018robust}                   
& Wi-Fi CSI
& \textcolor[RGB]{77,168,67}{\ding{51}}
& Feature Optimization
& \fullcirc
& Feature Generation
& \textcolor[RGB]{197,5,56}{\ding{55}}
& \textcolor[RGB]{77,168,67}{\ding{51}} \\
\hline
CSI4Free~\cite{bhat2024csi4free}           
& Wi-Fi CSI
& \textcolor[RGB]{77,168,67}{\ding{51}}
& Data Augmentation
& \emptycirc
& Real Signal Generation
& \textcolor[RGB]{197,5,56}{\ding{55}}
& \textcolor[RGB]{77,168,67}{\ding{51}} \\
\hline
WiARGAN~\cite{huang2021wiargan}                 
& Wi-Fi CSI
& \textcolor[RGB]{77,168,67}{\ding{51}}
& Data Augmentation
& \emptycirc
& Real Signal Generation
& \textcolor[RGB]{197,5,56}{\ding{55}}
& \textcolor[RGB]{77,168,67}{\ding{51}} \\
\hline
WTF-DCGAN~\cite{li2018wavelet}               
& Wi-Fi CSI
& \textcolor[RGB]{77,168,67}{\ding{51}}
& Data Balancing
& \emptycirc
& Feature Generation
& \textcolor[RGB]{197,5,56}{\ding{55}}
& \textcolor[RGB]{77,168,67}{\ding{51}} \\
\hline
WaveEar~\cite{xu2019waveear}                 
& Time-Frequency Characteristics, Reflection Characteristics
& \textcolor[RGB]{197,5,56}{\ding{55}}
& Feature Optimization
& \fullcirc
& Real Signal Generation, Environmental Reconstruction
& \textcolor[RGB]{197,5,56}{\ding{55}}
& \textcolor[RGB]{77,168,67}{\ding{51}} \\
\hline
SCL~\cite{liu2024wireless}                     
& Wi-Fi RSSI
& \textcolor[RGB]{77,168,67}{\ding{51}}
& Anomaly Detection
& \fullcirc
& Feature Generation, Environmental Reconstruction
& \textcolor[RGB]{197,5,56}{\ding{55}}
& \textcolor[RGB]{77,168,67}{\ding{51}} \\
\hline
RecTrack-GAN~\cite{belmonte2019recurrent}            
& Wi-Fi RSSI
& \textcolor[RGB]{77,168,67}{\ding{51}}
& Data Completion
& \fullcirc
& Environmental Reconstruction
& \textcolor[RGB]{197,5,56}{\ding{55}}
& \textcolor[RGB]{77,168,67}{\ding{51}} \\
\hline
G-HFD~\cite{wang2024generative}                   
& Wi-Fi CSI
& \textcolor[RGB]{77,168,67}{\ding{51}}
& Feature Optimization, Super-Resolution
& \fullcirc
& Feature Generation, Environmental Reconstruction
& \textcolor[RGB]{197,5,56}{\ding{55}}
& \textcolor[RGB]{77,168,67}{\ding{51}} \\
\hline
WiSIA~\cite{li2020wi}                   
& Wi-Fi RSSI
& \textcolor[RGB]{77,168,67}{\ding{51}}
& Image Generation, Super Resolution
& \fullcirc
& Feature Generation
& \textcolor[RGB]{197,5,56}{\ding{55}}
& \textcolor[RGB]{77,168,67}{\ding{51}} \\
\hline
GAN-LTE~\cite{serreli2024generative}                 
& LTE RSSI
& \textcolor[RGB]{77,168,67}{\ding{51}}
& Data Completion, Data Augmentation
& \emptycirc
& Feature Generation
& \textcolor[RGB]{197,5,56}{\ding{55}}
& \textcolor[RGB]{77,168,67}{\ding{51}} \\
\hline
Semi-supervised-VAE~\cite{chidlovskii2019semi}     
& WiFi-RSSI, Environmental Information
& \textcolor[RGB]{77,168,67}{\ding{51}}
& Data Repair, Feature Optimization
& \emptycirc
& Feature Generation, Environmental Reconstruction
& \textcolor[RGB]{197,5,56}{\ding{55}}
& \textcolor[RGB]{77,168,67}{\ding{51}} \\
\hline
VAE-LSTM~\cite{kim2023abnormal}                
& Wi-Fi CSI
& \textcolor[RGB]{77,168,67}{\ding{51}}
& Anomaly Detection
& \fullcirc
& Feature Generation
& \textcolor[RGB]{197,5,56}{\ding{55}}
& \textcolor[RGB]{197,5,56}{\ding{55}} \\
\hline
MoPoE-VAE~\cite{strohmayer2024through}               
& Wi-Fi CSI
& \textcolor[RGB]{77,168,67}{\ding{51}}
& Image Generation, Feature Optimization
& \fullcirc
& Feature Generation
& \textcolor[RGB]{77,168,67}{\ding{51}}
& \textcolor[RGB]{197,5,56}{\ding{55}} \\
\hline
VAWSS~\cite{zhang2024vawss}                   
& Wi-Fi CSI, Environmental Information, Ray Tracing
& \textcolor[RGB]{77,168,67}{\ding{51}}
& Data Completion
& \emptycirc
& Feature Generation
& \textcolor[RGB]{197,5,56}{\ding{55}}
& \textcolor[RGB]{77,168,67}{\ding{51}} \\
\hline
FiDo~\cite{chen2020fido}                    
& Wi-Fi CSI
& \textcolor[RGB]{77,168,67}{\ding{51}}
& Domain Generalization, Data Completion
& \emptycirc
& Feature Generation;
& \textcolor[RGB]{197,5,56}{\ding{55}}
& \textcolor[RGB]{77,168,67}{\ding{51}} \\
\hline
Fidora~\cite{chen2022fidora}                  
& Wi-Fi CSI
& \textcolor[RGB]{77,168,67}{\ding{51}}
& Data Completion
& \emptycirc
& Feature Generation
& \textcolor[RGB]{197,5,56}{\ding{55}}
& \textcolor[RGB]{77,168,67}{\ding{51}} \\
\bottomrule
\end{tabular}
\end{table*}

\begin{table*}[htbp]
\centering
\addtocounter{table}{-1}
\caption{Representative Wireless Sensing Applications Leveraging Generative AI (Continued)}
\label{tab:example-continued-2}

\small 
% 使用 p{固定宽度} 并在需要处加上 >{\raggedright} 或 >{\centering} 等控制对齐方式
\renewcommand\arraystretch{1.5}
\begin{tabular}{
  >{\centering\arraybackslash}m{2.2cm}|  % 1. System Name
  >{\centering\arraybackslash}m{2.3cm}|  
  %\raggedright
  % 2. Input (Description)
  >{\centering\arraybackslash}m{0.9cm}|    % 3. Input (Raw Signal)
  >{\centering\arraybackslash}m{2.3cm}|  % 4. Downstream Tasks (Description)
  >{\centering\arraybackslash}m{0.9cm}|   % 5. Downstream Tasks (Task types)
  >{\centering\arraybackslash}m{2.3cm}|  % 6. Generation's Target (Description)
  >{\centering\arraybackslash}m{0.9cm}|    % 7. Generation's Target (Multimodal)
  >{\centering\arraybackslash}m{1.4cm}    % 8. Conditional Generation
}
\hline
\multirow{2}{*}[-0.6em]{\centering \textbf{System Name}} 
& \multicolumn{2}{c|}{\textbf{Input}} 
& \multicolumn{2}{c|}{\textbf{Downstream Tasks}}
& \multicolumn{2}{c|}{\textbf{Generation's Target}}
& \multirow{2}{*}[-0.6em]{\centering \textbf{Condition}} \\
\cline{2-3} \cline{4-5} \cline{6-7}
& \centering \textbf{Description} 
& \centering \textbf{Raw Signal} 
& \centering \textbf{Description} 
& \centering \textbf{Task types} 
& \centering \textbf{Description} 
& \centering \textbf{Multi modal} 
& \\
\hline
EfficientFi~\cite{yang2022efficientfi}             
& Wi-Fi CSI
& \textcolor[RGB]{77,168,67}{\ding{51}}
& Feature Optimization
& \fullcirc
& Feature Generation
& \textcolor[RGB]{197,5,56}{\ding{55}}
& \textcolor[RGB]{197,5,56}{\ding{55}} \\
\hline
Delayed-Fusing~\cite{cominelli2023accurate}          
& Wi-Fi CSI
& \textcolor[RGB]{77,168,67}{\ding{51}}
& Anomaly Detection, Feature Optimization
& \fullcirc
& Feature Generation
& \textcolor[RGB]{197,5,56}{\ding{55}}
& \textcolor[RGB]{197,5,56}{\ding{55}} \\
\hline
RA-HDL~\cite{lin2023wi}                  
& Wi-Fi RSSI
& \textcolor[RGB]{77,168,67}{\ding{51}}
& Feature Optimization, Data Completion
& \fullcirc
& Feature Generation, Environmental Reconstruction
& \textcolor[RGB]{197,5,56}{\ding{55}}
& \textcolor[RGB]{77,168,67}{\ding{51}} \\
\hline
TriSense~\cite{khan2024trisense}                
& Wi-Fi CSI, Wi-Fi RSSI, Radar, RFID
& \textcolor[RGB]{77,168,67}{\ding{51}}
& Domain Generalization, Data Completion
& \halfcirc
& Cross-Modal Data Generation, Feature Generation
& \textcolor[RGB]{77,168,67}{\ding{51}}
& \textcolor[RGB]{77,168,67}{\ding{51}} \\
\hline
Rfidar~\cite{khan4824678rfidar}                   
& RFID, Radar
& \textcolor[RGB]{77,168,67}{\ding{51}}
& Data Completion, Feature Optimization
& \halfcirc
& Feature Generation, Cross-Modal Data Generation
& \textcolor[RGB]{77,168,67}{\ding{51}}
& \textcolor[RGB]{77,168,67}{\ding{51}} \\
\hline
WiGAN~\cite{jiang2020wigan}                   
& Wi-Fi CSI
& \textcolor[RGB]{77,168,67}{\ding{51}}
& Data Balancing, Feature Optimization
& \halfcirc
& Feature Generation
& \textcolor[RGB]{197,5,56}{\ding{55}}
& \textcolor[RGB]{77,168,67}{\ding{51}} \\
\hline
MCBAR~\cite{wang2021multimodal}                   
& Wi-Fi CSI
& \textcolor[RGB]{77,168,67}{\ding{51}}
& Data Completion, Domain Generalization
& \emptycirc
& Feature Generation
& \textcolor[RGB]{197,5,56}{\ding{55}}
& \textcolor[RGB]{77,168,67}{\ding{51}} \\
\hline
CAE~\cite{nirmal2020combating}                     
& Wi-Fi CSI
& \textcolor[RGB]{77,168,67}{\ding{51}}
& Feature Optimization, Image Generation
& \fullcirc
& Feature Generation
& \textcolor[RGB]{197,5,56}{\ding{55}}
& \textcolor[RGB]{77,168,67}{\ding{51}} \\
\hline
CrossGR~\cite{li2021crossgr}                 
& Wi-Fi CSI
& \textcolor[RGB]{77,168,67}{\ding{51}}
& Data Completion
& \emptycirc
& Real Signal Generation
& \textcolor[RGB]{197,5,56}{\ding{55}}
& \textcolor[RGB]{77,168,67}{\ding{51}} \\
\hline
Wi-CHAR~\cite{hao2024wi}                 
& Wi-Fi CSI
& \textcolor[RGB]{77,168,67}{\ding{51}}
& Feature Optimization
& \fullcirc
& Feature Generation, Real Signal Generation
& \textcolor[RGB]{197,5,56}{\ding{55}}
& \textcolor[RGB]{77,168,67}{\ding{51}} \\
\hline
WiFi2Radar~\cite{nirmal2024wifi2radar}              
& Wi-Fi CSI, Radar
& \textcolor[RGB]{77,168,67}{\ding{51}}
& Feature Optimization
& \halfcirc
& Feature Generation
& \textcolor[RGB]{197,5,56}{\ding{55}}
& \textcolor[RGB]{77,168,67}{\ding{51}} \\
\hline
RFBoost~\cite{hou2024rfboost}                 
& Time-Frequency Characteristics
& \textcolor[RGB]{197,5,56}{\ding{55}}
& Data Completion, Data Balancing
& \emptycirc
& Feature Generation, Network Optimization
& \textcolor[RGB]{197,5,56}{\ding{55}}
& \textcolor[RGB]{197,5,56}{\ding{55}} \\
\hline
TFSemantic~\cite{liao2024tfsemantic}             
& RFID, Radar
& \textcolor[RGB]{77,168,67}{\ding{51}}
& Data Balancing
& \emptycirc
& Feature Generation
& \textcolor[RGB]{197,5,56}{\ding{55}}
& \textcolor[RGB]{77,168,67}{\ding{51}} \\
\hline
CAE-MAS~\cite{raeis2024cae}                 
& Radar
& \textcolor[RGB]{77,168,67}{\ding{51}}
& Feature Optimization
& \fullcirc
& Feature Generation
& \textcolor[RGB]{197,5,56}{\ding{55}}
& \textcolor[RGB]{197,5,56}{\ding{55}} \\
\hline
CrossSense~\cite{zhang2018crosssense}              
& Wi-Fi CSI
& \textcolor[RGB]{77,168,67}{\ding{51}}
& Domain Generalization
& \emptycirc
& Network Optimization
& \textcolor[RGB]{197,5,56}{\ding{55}}
& \textcolor[RGB]{77,168,67}{\ding{51}} \\
\hline
LLM4CP~\cite{liu2024llm4cp}                  
& Wi-Fi CSI
& \textcolor[RGB]{77,168,67}{\ding{51}}
& Channel Prediction
& \fullcirc
& Real Signal Generation
& \textcolor[RGB]{197,5,56}{\ding{55}}
& \textcolor[RGB]{77,168,67}{\ding{51}} \\
\hline
Smartbond~\cite{karmakar2020smartbond}               
& CSI
& \textcolor[RGB]{77,168,67}{\ding{51}}
& Channel Prediction
& \fullcirc
& Network Optimization
& \textcolor[RGB]{197,5,56}{\ding{55}}
& \textcolor[RGB]{77,168,67}{\ding{51}} \\
\hline
SGAN~\cite{njima2021indoor}                    
& Wi-Fi RSSI
& \textcolor[RGB]{77,168,67}{\ding{51}}
& Data Augmentation
& \emptycirc
& Real Signal Generation, Environmental Reconstruction
& \textcolor[RGB]{197,5,56}{\ding{55}}
& \textcolor[RGB]{77,168,67}{\ding{51}} \\
\hline
GenAINet~\cite{zou2024genainet}                
& Wi-Fi CSI
& \textcolor[RGB]{77,168,67}{\ding{51}}
& Feature Optimization
& \emptycirc
& Network Optimization
& \textcolor[RGB]{77,168,67}{\ding{51}}
& \textcolor[RGB]{77,168,67}{\ding{51}} \\
\bottomrule
\end{tabular}
\end{table*}

The integration of GenAI into wireless sensing systems can be systematically understood through the lens of the sensing pipeline. A typical pipeline, as illustrated in \fig\ref{fig:pipeline-traditional}, begins with a physical-layer signal or extracted feature as input, which is processed by a discriminative model, such as a signal processing algorithm or task-specific deep learning network, and ultimately produces a high-level output such as localization, activity recognition, or environmental mapping.

Generative models reshape this pipeline by introducing flexible mechanisms for signal generation, representation learning, and cross-modal translation. The key distinction lies in the execution strategy: generative components can be incorporated as auxiliary modules to enhance existing discriminative models, as shown in \fig\ref{fig:pipeline-plugin}, or serve as end-to-end task solvers that directly map inputs to outputs, as depicted in \fig\ref{fig:pipeline-sovler}. Furthermore, generative models often incorporate conditional inputs, such as environmental layouts, device positions, or signal priors, which improve controllability, increase realism, and enhance generalization across scenarios. This leads to greater diversity in both input modalities and output representations compared to conventional sensing systems.

% To provide a structured view of these interactions, we analyze the generative sensing pipeline from four perspectives: the input signal modalities, the target outputs of the generative process, the execution strategy of the generative model, and whether conditional constraints are incorporated during the generation process. These four dimensions are reflected across a broad range of recent wireless sensing systems, as summarized in Table~\ref{tab:example}, which categorizes representative works based on their generative roles, signal inputs, tasks, and conditional structures.

To provide a structured view of these interactions, we analyze the generative sensing pipeline from four perspectives: the input signal modalities, the target outputs of the generative process, the execution strategy of the generative model, and whether conditional constraints are incorporated during generation. These dimensions are reflected across a broad range of recent wireless sensing systems, as summarized in Table~\ref{tab:example}, which categorizes representative works by their generative roles, input types, task objectives, and use of conditional inputs.

\subsection{Input Modalities}
In wireless sensing systems, the state of received wireless signals can be used as the inputs of the generative model. These signals may come from commercial wireless devices or Software-Defined Radios (SDRs). The following sections explore three primary signal types: received signal strength indicator (RSSI), CSI, and radar, which are widely used for relevant tasks. 
\subsubsection{RSSI}
RSSI is a general wireless signal strength measurement metric that can be directly obtained from most commercial wireless devices, such as Wi-Fi~\cite{10.1145/3055031.3055084, joshi2015decimeter}, LoRa~\cite{lam2017lora, lin2021sateloc}, and RFID~\cite{liu2011mining}, and is used to assess the spatial attenuation characteristics of signal propagation. Unlike other signal measurement methods, RSSI is highly influenced by environmental factors and typically follows propagation attenuation models, including path loss, shadowing, and multi-path models\cite{seidel1992914}, which describes the effects of absorption, reflection, scattering, and diffraction of obstacles on signal power. Compared to precise ranging technologies like Ultra-Wideband (UWB), the average RSSI may exhibit a monotonically decreasing relationship with distance on a large scale, making it suitable for coarse power-based ranging. However, due to the influence of random shadowing effects, this relationship is not stable on a relatively small scale. Additionally, RSSI is susceptible to multipath effects, causing fluctuations on the scale of the signal wavelength, which limits its applicability in fine-grained positioning.
\subsubsection{CSI}
CSI provides finer-grained wireless channel measurement capabilities compared to RSSI, capturing rich physical-layer information such as multi-path effects, phase variations, and path loss experienced by wireless signals during propagation~\cite{ma2019wifi}. The high-resolution CSI data reflects subtle environmental changes and exhibit high sensitivity to the spatiotemporal characteristics of the channel. Additionally, CSI demonstrates strong noise resistance and stability, as it not only relies on overall signal strength but also analyzes the independent characteristics of individual subcarriers, enabling more precise channel estimation in complex environments. CSI can further provide detailed signal propagation information across different frequencies and different antenna elements, making it suitable for more refined channel modeling and analysis. These characteristics make CSI a crucial foundation in the field of wireless sensing, offering extensive possibilities for intelligent signal analysis and environmental perception.
\subsubsection{Radar}
% 分别简介
Radar signals usually feature high time and spatial resolution, enabling precise perception capabilities in complex environments. For example, FMCW radar sends chirp signals\cite{mao2018aim}, and when they encounter a target and reflect back, the frequency of the received signal differs from the transmitted signal\cite{mao2016high, mao2017indoor}. By measuring this frequency difference, the distance and relative velocity of the target can be calculated. FMCW radar can improve distance resolution by increasing the frequency bandwidth, achieving high-precision human posture recognition, and is therefore widely applied in the field of wireless sensing~\cite{zhao2018through,zhao2018rf}.

\subsection{Generative Objectives}
%
% As wireless sensing tasks, generative artificial intelligence is widely applied in areas such as signal generation, feature extraction, environment reconstruction, and network optimization, thanks to its advantages in data augmentation, diversity enhancement, and pattern learning.
GenAI has found widespread use in wireless sensing tasks, including signal generation, feature extraction, environment reconstruction, and network optimization.

\subsubsection{Signal Generation}
Generative models have wide applications in signal generation for wireless sensing, including wireless signal synthesis, channel modeling, interference mitigation, and cross-modal signal generation. It can simulate real-world channel environments to generate realistic Wi-Fi, LoRa, or 5G signals, supporting data augmentation and testing for wireless sensing tasks\cite{bhat2024csi4free}, \cite{chen2023rf}, \cite{khan4824678rfidar}. By learning the underlying features of wireless signals, it optimizes channel modeling, improving the accuracy and robustness of environmental sensing. Cross-modal signal generation enhances data diversity, strengthening the domain generalization ability of wireless sensing models. Additionally, it can be used to enhance signal quality through denoising, completion, or amplification of weak signals, improving target detection, activity recognition, and scene reconstruction in complex environments\cite{serreli2024generative}, \cite{wang2021multimodal}. These advancements significantly enhance the precision, robustness, and adaptability of wireless sensing, driving the development of intelligent wireless perception.
\subsubsection{Feature Generation}
Feature generation using generative models could automatically generate meaningful information from complex wireless signals to support tasks such as object detection, activity recognition, and environmental sensing\cite{yang2022rethinking}, \cite{wu2023cddm}, \cite{liao2024tfsemantic}. Through self-supervised learning, GenAI can generate key features from unlabeled wireless signals, reducing the reliance on large amounts of labeled data and improving the adaptability of sensing systems. In cross-modal tasks, GenAI can fuse features from different types of signals, such as Wi-Fi signals combined with visual or audio signals, enhancing the multi-modal understanding of sensing models. Additionally, the advantages of GenAI in denoising and data completion enable it to generate clearer, more reliable features from noisy environments, improving the performance of sensing systems in complex environments and optimizing the accuracy and robustness of tasks like object detection, activity analysis, and scene reconstruction\cite{10279463}, \cite{chen2022fidora}, \cite{10.1145/3677525.3678663}.
\subsubsection{Environment Reconstruction}
Generative model has significant technical advantages in wireless sensing tasks with output objectives such as radio map generation and environment prediction. By learning the propagation characteristics of wireless signals, GenAI can construct high-precision radio environment models, reducing the need for actual measurements and improving network planning and optimization efficiency\cite{seong2019selective}. Additionally, GenAI can predict changes in the wireless environment, proactively identify potential sources of interference, and enhance the system’s adaptability and robustness. In environmental reconstruction, GenAI uses limited sensing data to infer and reconstruct the physical characteristics of the environment\cite{zhang2023rme}, providing a foundation for wireless sensing tasks like indoor positioning.

\subsection{Execution Strategies}

As illustrated in \fig\ref{fig:pipeline}, GenAI can be integrated into wireless sensing pipelines in two main ways: as an auxiliary plugin or as a standalone task solver. This integration has significantly transformed how wireless systems acquire, process, and interpret signal information. Unlike traditional pipelines that rely on feature extraction and discriminative models, generative approaches enable signal synthesis, domain adaptation, and robust representation learning, leading to more adaptive and generalizable solutions.

% GenAI has significantly transformed wireless sensing systems by redefining how information is acquired, processed, and interpreted. Unlike traditional pipelines that rely on feature extraction and discriminative models, generative approaches offer the ability to synthesize, complete, and translate signal data, enabling more adaptive and robust solutions. Depending on their integration within the sensing pipeline, generative models are typically employed in two ways: (1) as task solvers that directly perform core sensing functions, and (2) as auxiliary plugins that enhance existing models through data augmentation and domain adaptation.

% As illustrated in Fig.~\ref{fig:pipeline}, GenAI can be integrated into wireless sensing pipelines in two primary ways: as an auxiliary plugin or as a standalone task solver. This paradigm shift enables new capabilities beyond traditional discriminative pipelines—such as signal synthesis, domain adaptation, and robust representation learning—thereby improving system adaptability, generalization, and resilience in dynamic environments.

\subsubsection{Generative Models as Task Solvers}

When generative models serve as task solvers, the output of GenAI is often closely related to the sensing task and optimization objectives. As a primary technical support for wireless sensing, generative models directly contribute to the task output of wireless sensing.
\par
Concerning the task of wireless signal environment modeling, researchers investigate the generative models like GAN, VAE. Regarding environmental wireless signal modeling, research often focuses on tasks such as radio map prediction, super-resolution, and activity prediction. For instance, in \cite{seong2019selective}, researchers proposed an unsupervised learning-based WiFi fingerprinting system, where the coordinates of access points (APs) and fingerprint vectors were fed into a GAN model. The generator in the GAN learned the reference floor’s signal distribution and generated radio maps for other floors, while the discriminator improved the generator’s accuracy by distinguishing real data from generated data. Similarly, studies such as \cite{zhang2023rme} adopted a comparable approach named RME-GAN for radio map prediction. As shown in \fig\ref{fig:motivation-direct}, compared to traditional methods like MBI interpolation\cite{lee2012voronoi}, RBF interpolation\cite{krumm2003minimizing}, AE\cite{teganya2021deep}, DeepAE\cite{teganya2021deep}, RadioUnet\cite{levie2021radiounet}, generative approaches have achieved notable improvements. In \cite{li2020wi}, the WiSIA model is proposed to enhance the resolution of Wi-Fi imaging by leveraging cGANs. By using the coarse wavefront images generated from Wi-Fi signals as input, the cGAN model learns to generate refined segmentation masks. Through the adversarial learning mechanism, the generator aims to produce realistic segmentation masks, while the discriminator distinguishes between the generated and real segmentation masks. This GenAI approach effectively improves the resolution of Wi-Fi imaging. In contrast, \cite{strohmayer2024through} adopts a VAE model for Wi-Fi imaging. These works show the potential of generative models in wireless imaging field. 
\par
Due to the strong representational capabilities and data-driven nature, GenAI has been widely applied to feature extraction tasks in wireless sensing. In \cite{lin2023wi}, a VAE model is utilized to optimize the feature representation of Wi-Fi RSSI data. Through unsupervised learning, the VAE reconstructs input data while learning robust latent feature representations, with Kullback-Leibler (KL) divergence regularization incorporated to mitigate overfitting. In semi-supervised learning, the VAE generates pseudo-labels to further enhance feature quality. Ultimately, by employing dense skip connections for feature fusion, the VAE significantly improves the performance of indoor localization and navigation.
%%许老师建议不要在这里讨论通信相关内容
%From a communication optimization perspective, \cite{wu2023cddm} employs a Diffusion model to learn the distributional characteristics of channel input signals. By modeling the noise distribution and progressively denoising the signals, the approach enhances communication performance under low SNR conditions. 
Additionally, generative models have been applied to feature extraction across different input modalities, including Wi-Fi CSI \cite{jiang2020wigan}\cite{nirmal2020combating}, Radar \cite{khan4824678rfidar}\cite{khan2024trisense}, and RFID \cite{khan4824678rfidar}. Based on the feature extraction capacity of the generative model, researchers utilize it for anomaly detection such as activity anomaly detection, wireless signal anomaly detections\cite{kim2023abnormal}\cite{toma2020deep}\cite{cominelli2023accurate}.
\par
Moreover, GenAI has also been employed for channel prediction in wireless sensing tasks. Channel prediction aims to forecast the characteristics of the wireless channel at a future time or location based on historical or current wireless signal information, such as CSI. Since wireless channels are affected by factors like multipath effects, moving obstacles, and environmental changes, channel prediction can help improve the reliability and efficiency of wireless communication systems, while also supporting wireless sensing tasks such as target tracking and environmental awareness. Various GenAI architectures have been widely applied to this task. \cite{chi2024rf} introduces a Diffusion-based approach for 5G downlink channel prediction, where the model learns the relationship between uplink and downlink channels to enable channel prediction under Frequency Division Duplex (FDD) transmission strategies. \cite{balevi2021wideband} presents a GAN-based method for channel prediction in low SNR environments. Additionally, \cite{baur2022variational} employs a VAE to learn channel characteristics, effectively enhancing the accuracy of channel prediction.
\subsubsection{Generative Models as Plugins}
In addition to directly participating in wireless sensing tasks, GenAI can serve as a supporting module by providing upstream task support for wireless sensing through model outputs, including data augmentation, data completion, and data balancing. In this process, generative models first learn the latent probability distribution from a large volume of real data, capturing key features and patterns to ensure that the generated samples are statistically similar to the original data. Once the data distribution is learned, the model can sample from it to synthesize new data. For example, GANs\cite{mirza2014conditional} generate samples that are indistinguishable from real data through adversarial training, VAEs\cite{sohn2015learning} learn latent space representations and sample from them to generate new data, while Diffusion Models\cite{dhariwal2021diffusion} progressively add noise and learn the denoising process to produce high-quality data.
\par
Employing GenAI as plugins for wireless sensing is particularly promising to address the challenges of sample scarcity\cite{jiang2020wigan}, and model generalization\cite{zinys2021domain} significantly affect system performance and practicality. 
\par
First, sample scarcity is mainly due to the limited availability of labeled data. The collection of wireless sensing data often involves complex experimental environments, high costs, and strict privacy protection requirements, which make it difficult to obtain high-quality training data. This data shortage limits the training effectiveness of deep learning models and makes it hard for them to learn comprehensive feature distributions. To solve the  data sample scarcity problem in wireless sensing, researchers use generative model for the objectives of data argumentation\cite{jiang2020wigan}\cite{wang2025generative}\cite{mao2024wi}\cite{chi2024rf}\cite{zhao2023nerf2}\cite{yang2022rethinking}\cite{chen2023rf}\cite{li2024diffgait}, data completion\cite{vishwakarma2021gan}\cite{liao2024tfsemantic}\cite{seong2019selective}\cite{zhang2024vawss}\cite{chen2022fidora}, and data balance\cite{hou2024rfboost}\cite{H6}. By leveraging the data generation capabilities of GenAI, numerous widely used wireless sensing tasks have achieved significant performance improvements. In the field of localization, \cite{chan2024learning} employs a GAN for WiFi fingerprint inpainting, utilizing deep learning models to capture correlations both among and within WiFi APs. This approach enables the completion of missing WiFi fingerprint data in unmeasured areas, thereby enhancing indoor localization accuracy. Similarly, \cite{junoh2023enhancing} utilizes GANs to generate wireless signal fingerprint data, incorporating Bluetooth Low Energy (BLE) beacons to augment the training dataset. This method improves localization precision and stability while reducing measurement time and manual effort. Moreover, \cite{cui2023more} introduces LocGAN, a semi-supervised deep generative model based on GANs, which leverages a small number of labeled WiFi fingerprints along with a large volume of unlabeled data to generate virtual fingerprints (VFPs), thereby reducing site survey costs and improving localization accuracy. Additionally, \cite{njima2021gan} combines GANs with a limited amount of labeled and unlabeled data to generate synthetic fingerprint data, which is then mixed with real data for training Deep Neural Network (DNN), ultimately enhancing indoor localization accuracy. Furthermore, \cite{L3} proposes an Amplitude Feature-based Deep Convolutional GAN (AF-DCGAN), which leverages GANs to generate additional amplitude feature maps, expanding the initial fingerprint database and improving indoor localization accuracy without increasing manual data collection efforts.
\par
Model generalization is another challenge faced by wireless sensing tasks. The propagation characteristics of wireless signals are strongly influenced by environmental factors, which makes it difficult for models trained in one environment to be directly transferred to a new one. Furthermore, dynamic changes in channel conditions, hardware differences, and external interference exacerbate the environmental dependency of the models, making it difficult to guarantee the stability and reliability across different scenarios. Although deep learning models can achieve high accuracy on specific datasets, their performance often significantly declines under different deployment conditions, especially when signal propagation uncertainties, hardware biases between devices, and data distribution shifts affect the model’s generalization ability. Therefore, ensuring the robustness and adaptability of models in diverse environments has become a key focus of research.
In \cite{wang2021multimodal}, researchers introduce a multi-modal generator and a cross-domain transformation generator to enhance model generalization. In \cite{han2020deep}, researchers have utilized GANs to generate synthetic samples to expand the training set. In conjunction with Multi-kernel Maximum Mean Discrepancy (MK-MMD) for domain adaptation, they reduce the distribution discrepancy between domains, thereby achieving high-precision gesture recognition and domain generalization across different environments.
\par
In summary, the application of GenAI provides new optimization solutions and performance enhancement perspectives for wireless sensing tasks.

\subsection{Conditional Constraints}

Conditional generation has become a core capability in modern generative modeling, enabling the integration of auxiliary information such as class labels, signal characteristics, or environmental context into the generation process. This conditioning mechanism improves controllability, enhances sample relevance, and boosts generalization. Notably, it is a model-agnostic design and has been adopted in Conditional GANs~\cite{mirza2014conditional}, Conditional VAEs~\cite{sohn2015learning}, and Conditional Diffusion Models~\cite{dhariwal2021diffusion}.

In wireless sensing tasks, the introduction of conditional inputs is crucial, as the propagation characteristics of wireless signals are influenced by various environmental factors, such as multipath effects, fading, interference, and dynamically changing user behavior. Conditional generative models offer an effective solution by incorporating auxiliary information into the data generation process, thereby enhancing model adaptability, improving controllability, and increasing robustness under diverse scenarios.

Several representative cases illustrate how conditional inputs are used in wireless sensing to guide generative models. One typical case involves using activity class labels to indicate specific human activity types such as walking, sitting, and standing. In these tasks, conditional generative models, including diffusion-based architectures, incorporate class information to synthesize signal data that aligns with target behaviors, thereby improving generation accuracy and interpretability~\cite{10817390, 10824863}.

Another common form of conditioning uses signal feature labels to describe key attributes of wireless data, including signal strength, frequency variation, and multipath effects. These descriptors help the generative model better capture physical characteristics and reproduce data patterns more faithfully~\cite{huang2023diffar, chi2024rf, bian2024wi}.

Conditional generation can also benefit from environmental priors such as room layout, antenna configuration, and known user behavior patterns. Providing such spatial or contextual information allows the model to generate data that is more consistent with real-world propagation scenarios~\cite{qiu2023irdm, 10327032, liu2025wifi}. For instance, knowledge of different room geometries or antenna positions helps generative models adjust their output distribution to match varying signal environments.

These cases illustrate the practical value of conditional generative models in wireless sensing. By leveraging contextual information, such models can enhance the realism, diversity, and task alignment of generated data, ultimately contributing to more robust and adaptive wireless sensing systems.

\section{Generative AI Techniques for Wireless Sensing}
\label{sec:generative}

Recent advances in generative modeling have led to a diverse set of architectures with varying theoretical foundations and operational mechanisms. From early statistical models to deep neural generators, these techniques differ in how they learn data distributions, generate samples, and handle conditioning or multimodal fusion. In the context of wireless sensing, where signal diversity, environmental variability, and task-specific requirements are prevalent, understanding the strengths and limitations of each generative architecture becomes crucial.

\begin{figure*}[t]
     \centering
     \includegraphics[width=1\linewidth]{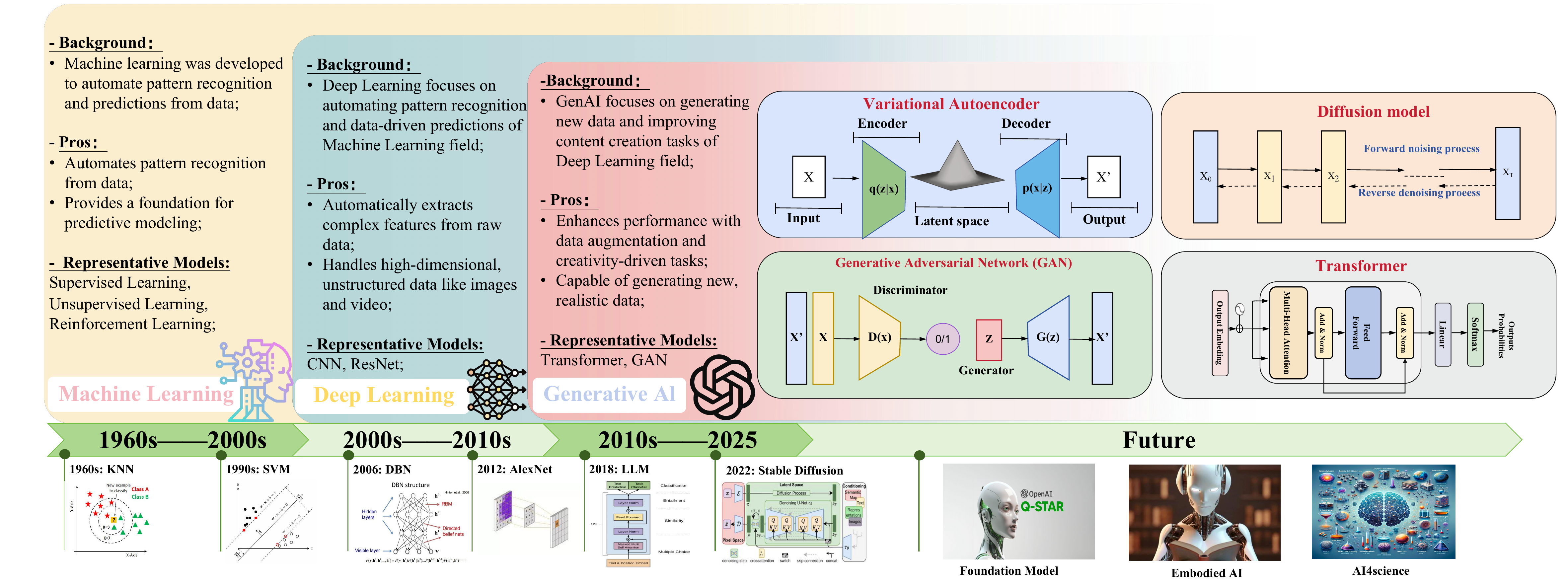}
     \caption{From traditional machine learning to GenAI: key milestones and representative generative model architectures.}
     \label{fig:generative}
 \end{figure*}

As shown in \fig\ref{fig:generative}, the evolution of artificial intelligence techniques has progressed from traditional machine learning models, which are limited in complexity, to deep learning models capable of handling large-scale, high-dimensional data, and ultimately to generative AI techniques, which excel at data generation tasks. This evolution reflects an increasing capability to model and generate data, which is critical for applications like wireless sensing where complex, high-dimensional signals need to be modeled, reconstructed, or generated. 

This section offers a model-centric perspective on GenAI, reviewing the core principles, representative structures, and practical applications of five mainstream generative approaches—Transformers, GANs, VAEs, Diffusion Models, and Large Language Models (LLMs). For each technique, we summarize its core mechanisms, highlight how it maps to wireless sensing tasks such as signal reconstruction, channel modeling, and activity recognition, and discuss practical trade-offs in terms of training stability, output quality, and adaptability to wireless data modalities.

\begin{table*}[p]
\centering
\caption{Generative Model Architectures and Their Applications in Wireless Sensing}
\label{tab:table_part1_nocol}
\small
\renewcommand\arraystretch{1.5}

% We now have 5 columns total (instead of 6).
\begin{tabular}{
  >{\centering\arraybackslash}m{2.2cm}|  % System Name
  >{\centering\arraybackslash}m{1.5cm}|  % Basic Structure
  >{\centering\arraybackslash}m{3.6cm}|  % Pros and Cons
  >{\centering\arraybackslash}m{3.4cm}|  % Specific Methods
  >{\centering\arraybackslash}m{4.0cm}   % Research Significance
}
\hline
\textbf{System Classification} 
& \textbf{Basic Structure} 
& \textbf{Pros and Cons} 
& \textbf{Specific Structure} 
& \textbf{Related Work} \\
\hline
%================================================
% MODEL-BASED (5 rows)

%================================================
% DATA-DRIVEN (NON-AUTOREGRESSIVE) => PART 1
  \multirow{11}{*}[-24em]{\centering Non-AR}
  & \multirow{4}{*}[-5em]{\centering Diffusion}
  & \multirow{4}{*}[-0.2em]{
    \begin{tabular}[c]{@{}l@{}}
    \textbf{Pros:}\\
    1. High-Quality Output\\
    2. Stable Training\\
    3. Flexible Architecture\\
    4. Robustness to Noise\\
    \textbf{Cons:}\\
    1. Slow Generation\\
    2. Parameter Sensitivity
    \end{tabular}
  }
  & \multirow{5}{*}[-0.5em]{\centering cDiffusion\cite{dhariwal2021diffusion}}
  & G-HFD\cite{wang2024generative} \\
  \cline{5-5}
  &
  &
  &
  & DiffAR\cite{huang2023diffar}\\
  \cline{5-5}
  &
  &
  &
  &M. Xu et al.\cite{10817390}\\
  \cline{5-5}
  &
  &
  &
  & RF-ACCLDM\cite{10824863}\\
  \cline{5-5}
  &
  &
  &
  &D. Guo et al.\cite{10327032}\\
  \cline{4-5}
  &
  &
  &\multirow{1}{*}[-0.2em]{\centering Latent Diffusion\cite{rombach2022high}}
  &DiffGait\cite{li2024diffgait}\\
  \cline{4-5}
  &
  &
  &\multirow{1}{*}[-0.2em]{\centering \makecell{TF-Diffusion}\cite{chi2024rf}}
  &RF-Diffusion\cite{chi2024rf}\\
  \cline{4-5}
  &
  &
  &\multirow{3}{*}[-0.2em]{\centering DDPM\cite{sohl2015deep}}
  &Diffradar\cite{10448324}\\
  \cline{5-5}
  &
  &
  &
  &AirECG\cite{10.1145/3678550}\\
  \cline{5-5}
  &
  &
  &
  &X. Luo et al.\cite{10843401}\\
  \cline{2-5}
  &\multirow{4}{*}[-15em]{\centering GAN}
  & \multirow{4}{*}[-8em]{
    \begin{tabular}[c]{@{}l@{}}
    \textbf{Pros:}\\
    1. High-Quality Output\\
    2. Efficient Generation\\
    3. Unsupervised Learning\\
    4. Model Customization\\
    \textbf{Cons:}\\
    1. Mode Collapse\\
    2. Hyperparameter \\Sensitivity\\
    3. High Resource Demand
    \end{tabular}}
  &\multirow{4}{*}[-0.5em]{\centering Cycle-GAN\cite{zhu2017unpaired}}
  &MaP-SGAN\cite{H14}\\
  \cline{5-5}
  &
  &
  &
  &Wi-Cro\cite{mao2024wi}\\
  \cline{5-5}
  &
  &
  &
  &C. Li et al.\cite{li2020wi}\\
  \cline{5-5}
  &
  &
  &
  &Wi-Diag\cite{10210319}\\
  \cline{4-5}
  &
  &
  &\multirow{6}{*}[-0.5em]{\centering WGAN\cite{arjovsky2017wasserstein}}
  &extendGAN+\cite{s23094402}\\
  \cline{5-5}
  &
  &
  &
  &R. Gopikrishna et al.\cite{10017520}\\
  \cline{5-5}
  &
  &
  &
  &LESS\cite{10011166}\\
  \cline{5-5}
  &
  &
  &
  &H. M. Nazmul et al.\cite{hasan2023wasserstein}\\
  \cline{5-5}
  &
  &
  &
  &L. Qu et al.\cite{9554556}\\
  \cline{5-5}
  &
  &
  &
  &F. Alharbi\cite{9206624}\\
  \cline{4-5}
  &
  &
  &\multirow{6}{*}[-0.5em]{\centering DCGAN\cite{radford2015unsupervised}}
  &AF-DCGAN\cite{8891678}\\
  \cline{5-5}
  &
  &
  &
  &Q. Li et al.\cite{li2018wavelet}\\
  \cline{5-5}
  &
  &
  &
  &WiGAN\cite{jiang2020wigan}\\
  \cline{5-5}
  &
  &
  &
  &C. Lim et al.\cite{9621134}\\
  \cline{5-5}
  &
  &
  &
  &H. Wu et al.\cite{10221373}\\
  \cline{5-5}
  &
  &
  &
  &Y. Lei et al. \cite{sym12091565}\\
  \cline{4-5}
  &
  &
  &\multirow{6}{*}[-0.5em]{\centering cGAN\cite{mirza2014conditional}}
  &CSI4Free\cite{bhat2024csi4free}\\
  \cline{5-5}
  &
  &
  &
  &S. A. Junoh et al.\cite{10443392}\\
  \cline{5-5}
  &
  &
  &
  &RME-GAN\cite{RME-GAN}\\
  \cline{5-5}
  &
  &
  &
  &S. A. Junoh et al.\cite{L9}\\
  \cline{5-5}
  &
  &
  &
  &Y. Zhang et al.\cite{10758412}\\
  \cline{5-5}
  &
  &
  &
  & J. Boulis et al.\cite{10.1145/3474717.3486807}\\
  \cline{2-5}
   &\multirow{11}{*}[-0.2em]{\centering VAE}
  & \multirow{10}{*}[-0.5em]{
    \begin{tabular}[c]{@{}l@{}}
    \textbf{Pros:}\\
     1. Probabilistic Framework\\
    2. Stable Training\\
    3. Smooth Latent Space\\
    \textbf{Cons:}\\
    1. Lower Output Quality\\
    2. Blurry Outputs\\
    3. Balancing Loss Terms\\
    4. Less Diverse Outputs
    \end{tabular}}
  &\multirow{2}{*}[-0.5em]{\centering$\beta$-VAE\cite{higgins2017beta}}
  &R. Yuan et al.\cite{10279463}\\
  \cline{5-5}
  &
  &
  &
  &M. Awais et al.\cite{10463503}\\
  \cline{4-5}
  &
  &
  &\multirow{6}{*}[-0.5em]{\centering cVAE\cite{sohn2015learning}}
  &DeepRssI\cite{10526268}\\
  \cline{5-5}
  &
  &
  &
  &Z. Lyu et al.\cite{10485223}\\
  \cline{5-5}
  &
  &
  &
  &Fidora\cite{chen2022fidora}\\
  \cline{5-5}
  &
  &
  &
  &W. Qian et al.\cite{9412651}\\
  \cline{5-5}
  &
  &
  &
  &J. Zhang et al.\cite{10571187}\\
  \cline{5-5}
  &
  &
  &
  &W. Qian et al.\cite{QIAN2021228}\\
  \cline{4-5}
  &
  &
  &\multirow{2}{*}[-0.5em]{\centering VQ-VAE\cite{van2017neural}}
  &V. Lafontaine et al.\cite{10.1145/3677525.3678663}\\
  \cline{5-5}
  &
  &
  &
  &S. K. Kompella et al.\cite{10773675}\\
  \cline{1-5}
\end{tabular}
\end{table*}
\begin{table*}[t]
\addtocounter{table}{-1}
\centering
\caption{Generative Model Architectures and Their Applications in Wireless Sensing (Continued)}
\label{tab:table_part2_nocol}
\small
\renewcommand\arraystretch{1.5}
\begin{tabular}{
   >{\centering\arraybackslash}m{2.2cm}|  % System Name
  >{\centering\arraybackslash}m{1.5cm}|  % Basic Structure
  >{\centering\arraybackslash}m{3.6cm}|  % Pros and Cons
  >{\centering\arraybackslash}m{3.4cm}|  % Specific Methods
  >{\centering\arraybackslash}m{4.0cm}   % Research Significance
}
\hline
\textbf{System Classification} 
& \textbf{Basic Structure} 
& \textbf{Pros and Cons} 
& \textbf{Specific Structure} 
& \textbf{Related Work} \\
\hline
  \multirow{11}{*}[-1em]{\centering Autoregressive}
  & \multirow{4}{*}[-7em]{\centering Transformer}
  & \multirow{5}{*}[0.2em]{
    \begin{tabular}[c]{@{}l@{}}
    \textbf{Pros:}\\
    1. Context Awareness\\
    2. Pretraining\\
    3. Flexible Input \\Representation\\
    4. Multimodal Learning\\
    \textbf{Cons:}\\
    1. High Computational \\Cost\\
    2. Data Dependency\\
    3. Training Instability\\
    \end{tabular}
  }
  & \multirow{7}{*}[-0.05em]{\centering ViT\cite{dosovitskiy2020image}}
  & F. Luo et al.\cite{luo2024vision} \\
  \cline{5-5}
  &
  &
  &
  &VTIL\cite{zhou2024vtil}\\
  \cline{5-5}
  &
  &
  &
  &T. Kim et al.\cite{10.1145/3632366.3632374}\\
  \cline{5-5}
  &
  &
  &
  &Witransformer\cite{yang2023witransformer}\\
  \cline{5-5}
  &
  &
  &
  &S. Masrur\cite{masrur2025transforming}\\
  \cline{5-5}
  &
  &
  &
  &MetaFi++\cite{zhou2023metafi++}\\
  \cline{4-5}
  &
  &
  &\multirow{4}{*}[-0.05em]{\centering LLM\cite{radford2018improving}}
  &LLMCount\cite{li2024llmcount}\\
  \cline{5-5}
  &
  &
  &
  &A. Kalita et al.\cite{kalita2024large}\\
  \cline{5-5}
  &
  &
  &
  &S. Dai et al.\cite{dai2024advancing}\\
  \cline{5-5}
  &
  &
  &
  &TIPS\cite{zhang2022tips}\\
  \cline{4-5}
  &
  &
  &\multirow{4}{*}[-0.0001em]{\centering Vanilla Transformer\cite{bahdanau2014neural}}
  &WiFiMod\cite{10.1145/3460112.3471951}\\
  \cline{5-5}
  &
  &
  &
  &A. Raza et al.\cite{raza2021lightweight}\\
  \cline{5-5}
  &
  &
  &
  &Y. Shavit\cite{shavit2021boosting}\\
  \cline{1-5}
\end{tabular}
\end{table*}

% \subsection{Physical Model}

\subsection{Generative Transformer}
Transformers have demonstrated strong potential in wireless sensing, particularly in gesture recognition and gait detection. Their advantage lies in the ability to capture long-range dependencies of wireless signals through self-attention and to learn feature patterns across different subspaces via multi-head attention, thereby distinguishing fine-grained motion characteristics. Recent studies have shown that Vision Transformer architectures can significantly improve the accuracy of Wi-Fi CSI-based human activity and gesture classification\cite{luo2024vision}, while another Transformer-based approach achieved nearly 99\% accuracy in gait recognition and person identification by jointly modeling CSI amplitude and phase \cite{ avola2025transformer}.\par
The Transformer follows a standard encoder–decoder framework. The process begins by mapping input wireless signal sequences such as CSI subcarrier magnitudes into embedding vectors, with positional encodings added to preserve temporal order. Within the encoder, self-attention computes the relationships among all time steps in the sequence, capturing long-range dependencies by generating Query, Key, and Value matrices, computing similarity scores, and aggregating weighted representations\cite{vaswani2017attention}. In practice, multi-head attention performs attention in parallel across different subspaces, enabling simultaneous learning of temporal fluctuations and frequency-domain features\cite{devlin2019bert}. After feed-forward layers and normalization, the encoder produces contextual representations, while the decoder integrates historical information and conditional inputs to generate outputs. This mechanism allows effective modeling of both long-term CSI dynamics and fine-grained motion differences\cite{dosovitskiy2020image}.\par
These architectural strengths explain why Transformers excel in wireless sensing tasks. In gesture recognition, self-attention establishes global connections among CSI variations across time, distinguishing highly similar action patterns. In gait detection, the combination of positional encoding and long-range modeling captures dynamic patterns of gait cycles. Compared with convolutional or recurrent models, Transformers provide parallel modeling capability and cross-temporal dependency capture, resulting in stronger robustness and generalization.\par
Nevertheless, Transformers also face limitations in wireless sensing. Their performance is strongly dependent on large annotated datasets, which are expensive to collect and label in this domain. Scaling law research has shown that Transformer accuracy follows predictable power-law improvements with model and dataset size\cite{kaplan2020scaling}, but insufficient data leads to overfitting. Furthermore, Transformers have high parameter counts and computational complexity; large models such as GPT-3\cite{brown2020language} and PaLM\cite{chowdhery2023palm} rely on massive compute resources, which are impractical for real-time deployment in wireless sensing. Finally, their generalization under low-sample conditions remains limited, requiring self-supervised or generative pretraining strategies to improve performance in tasks such as fall detection.\par

In summary, the operational principles of Transformer naturally with the spatio-temporal modeling requirements of wireless sensing, providing distinctive advantages in gesture recognition and gait detection. However, their reliance on computational resources and large-scale datasets poses challenges for practical deployment, motivating future research on lightweight architectures and cross-task pretraining to advance their applicability in wireless sensing.

\subsection{Generative Adversarial Network (GAN)}

In wireless sensing, GAN have demonstrated strong potential, particularly in the tasks of data augmentation and localization. Because the collection and annotation of wireless data are costly, datasets are often limited in size, making it difficult for models to generalize across complex environments. GANs alleviate this problem by learning to generate synthetic samples that closely resemble the distribution of real CSI or RSSI signals. For instance, conditional GANs have been used to expand Wi-Fi fingerprint databases, reducing site-survey efforts while maintaining localization accuracy\cite{ witteveen2019comparison}. Similarly, CycleGAN-based methods have enabled cross-domain fingerprint translation, allowing localization models to remain robust when deployed in new environments\cite{ chan2018surface}. These applications demonstrate that GANs are particularly effective in addressing data scarcity and domain shift challenges.
\par

From an operational perspective, GANs consist of a generator and a discriminator trained in a minimax game. The generator maps random noise or conditional variables into synthetic samples, while the discriminator attempts to distinguish between real and fake data. Over time, the generator improves its outputs, gradually approximating the true data distribution\cite{ goodfellow2020generative}. To stabilize training, Wasserstein GAN introduces the Earth-Mover distance in place of Jensen–Shannon divergence\cite{ adler2018banach}, while CycleGAN employs cycle-consistency loss to enable unpaired data translation, which is highly valuable for cross-domain wireless sensing\cite{ zhu2017unpaired}.
\par

These mechanisms align closely with the requirements of wireless sensing. In data augmentation, GANs can generate diverse CSI or RSSI samples to mitigate class imbalance and improve the robustness of activity recognition or fall detection models. In localization, GANs can synthesize virtual fingerprints to reduce the cost of data collection, while cross-domain GANs can adapt fingerprints across different environments or devices, enhancing the generalization of localization models. Furthermore, conditional GANs can incorporate side information such as location or activity labels, producing task-specific synthetic samples that further enhance performance.
\par

Moreover, GANs also face important limitations. Their training process is often unstable, and the adversarial game can lead to mode collapse, where the generator produces samples with limited diversity\cite{ gulrajani2017improved}\cite{ mescheder2018training}. In wireless sensing scenarios with already limited data, this issue can reduce the effectiveness of augmentation. GANs are also highly sensitive to hyperparameter settings and architectural choices, requiring careful design to maintain quality. Moreover, their computational demands are significant, which limits their suitability for real-time wireless sensing applications.
\par

Recent studies have explored directions to make GANs more practical in wireless sensing. The CSI4Free framework employs a cWGAN to generate high-fidelity mmWave CSI samples, significantly improving the generalization of pose classification models\cite{ bhat2024csi4free}. In localization, synthetic fingerprints generated by GANs have been shown to maintain accuracy while reducing real data requirements by up to 90\%\cite{ nabati2020using}. Moreover, Wi-Fi Semi-Supervised GANs  have demonstrated notable improvements in positioning accuracy by combining label information with generative training\cite{ yoo2024wi}. These findings suggest that with stabilized training objectives, lightweight designs, and the integration of wireless propagation priors, GANs are moving toward more practical, efficient, and accurate deployment in wireless sensing.

\subsection{Variational Autoencoder (VAE)}
VAEs have been widely applied in wireless sensing, particularly in anomaly detection, environment reconstruction, and data augmentation. Compared with GANs, VAEs offer the advantage of explicit probabilistic generative modeling, enabling the learning of latent distributions of input data and using reconstruction error as a criterion for modeling and detection. For example, in Wi-Fi CSI anomaly detection, VAEs effectively distinguish normal and abnormal signals by leveraging reconstruction error, supporting applications such as fall detection and intrusion detection\cite{ kim2023abnormal}. VAEs have also been applied in environment reconstruction, learning the latent relationship between CSI and spatial layouts to generate structural features for indoor layout modeling and imaging\cite{strohmayer2024through}. In data augmentation, studies have demonstrated that VAEs can generate synthetic CSI samples across diverse environments to improve the robustness of activity recognition models\cite{aparna2024human}.\par
The operational principle of VAEs relies on an encoder–decoder architecture. Wireless signal inputs are first mapped by the encoder into distribution parameters of the latent space, from which latent variables are sampled using the reparameterization trick. The decoder then reconstructs the input signal from these latent variables\cite{kingma2013auto}. By maximizing the evidence lower bound, VAEs balance generation fidelity with latent distribution modeling. This mechanism is particularly suitable for reconstruction-based wireless sensing tasks, as it naturally uses reconstruction error for anomaly detection and captures latent distribution features for cross-environment generalization.\par
Compared with other generative models, VAEs in wireless sensing offer the advantages of probabilistic interpretability and stable training. They generate diverse and distribution-consistent synthetic samples, providing reliable tools for data augmentation and environment modeling. However, VAE-generated samples are often blurry due to the KL-divergence regularization in their objective, making them less sharp than GAN outputs in terms of visual or signal detail\cite{ kingma2013auto}\cite{rezende2014stochastic}. Moreover, in modeling high-dimensional complex wireless signals, the expressive power of the latent space may be insufficient, limiting scalability in large-scale wireless sensing tasks\cite{higgins2017beta}.\par
Recent research has introduced improvements to expand VAEs toward broader applications in wireless sensing. Beta-VAE enhances disentanglement in the latent space, facilitating independent representation of different motion or spatial features\cite{ higgins2017beta}. Hierarchical VAEs increase generation capability by stacking multiple latent layers\cite{zhao2017towards}. Moreover, integrating VAEs with physical priors has been shown to advance their use in wireless imaging and environment modeling\cite{cao2025reconstructing}. These explorations indicate that VAEs are becoming a crucial generative tool in wireless sensing, balancing stability with probabilistic interpretability.\par

\subsection{Diffusion Model}
Diffusion models have recently emerged as a promising generative approach in wireless sensing, particularly in tasks such as environment reconstruction, gait detection, and data augmentation. Compared with GANs or VAEs, diffusion models provide superior stability and sample diversity. For example,\cite{liao2024view} demonstrates that diffusion-based gait recognition can leverage iterative denoising to recover fine-grained radar reflections, significantly improving recognition accuracy under cross-environment conditions. In environment reconstruction, diffusion-based radio map generation has been shown to synthesize high-resolution maps from sparse samples, outperforming interpolation and GAN-based baselines \cite{luo2025denoising}. Similarly, in data augmentation, conditioned diffusion models such as\cite{wang2025generative} enrich Wi-Fi CSI datasets by generating diverse synthetic samples, achieving up to 70% performance improvements in detection tasks.\par
The operational principle of diffusion models differs from that of other generative frameworks. In the forward process, Gaussian noise is gradually added to the data, transforming complex distributions into nearly isotropic noise; the reverse process, parameterized by neural networks, learns to progressively denoise and reconstruct the original distribution\cite{ho2020denoising}. Score-based generalizations formalize diffusion as a pair of stochastic differential equations, describing diffusion from data to noise and reverse sampling from noise to data \cite{song2020score}. This iterative denoising mechanism aligns naturally with wireless sensing, where signals are often corrupted by fading and noise, making diffusion theoretically well-suited for signal recovery and reconstruction tasks.\par
In wireless sensing applications, diffusion models offer multiple advantages. Their ability to generate diverse samples alleviates the mode collapse problem common in GANs, enabling better coverage of cross-domain CSI distributions. This property has proven particularly useful for gait detection and activity recognition, where diverse motion patterns must be captured to ensure robustness across environments. Additionally, their stepwise denoising process provides a stable reconstruction pathway, which has been leveraged effectively in radio map generation and indoor environment imaging.\par
Nonetheless, diffusion models also face limitations. Chief among these is their inefficiency at inference, since sampling typically requires dozens or even hundreds of iterations, making them unsuitable for real-time low-latency sensing. Although accelerated methods such as DDIM\cite{song2020denoising} reduce the number of steps, diffusion still lags behind one-shot generative models like GANs. Moreover, the training and inference of diffusion models are computationally demanding, restricting deployment on resource-constrained edge devices. Finally, their performance in low-data regimes remains an open challenge, as wireless sensing datasets are often limited.\par
Recent studies point to promising directions for advancing the adoption of diffusion models in wireless sensing. For instance, \cite{chi2024rf} introduces time–frequency domain diffusion to generate high-quality Wi-Fi and FMCW signals, achieving notable improvements in channel prediction and radar imaging. \cite{wang2025generative} leverages conditioned diffusion to augment limited CSI datasets, yielding substantial performance gains in detection. \cite{luo2025denoising} integrates conditional control into the diffusion process to produce high-fidelity radio maps under sparse sampling conditions. Together, these efforts suggest that with efficient sampling algorithms, conditional mechanisms, and physics-informed priors, diffusion models are poised to become a central tool in data augmentation, channel prediction, and environment modeling for wireless sensing.

\subsection{Large-language Model (LLM)}
LLMs have recently achieved groundbreaking success in artificial intelligence and are now being explored in wireless sensing tasks. Unlike traditional methods that design task-specific architectures, LLMs offer cross-task generalization, enabling a single framework to handle diverse sensing objectives such as fall detection, activity recognition, and localization. This advantage stems from their large-scale pretraining and in-context learning capabilities, which allow LLMs to adapt to new tasks with few or even zero examples. Recent work, for instance, proposed tokenizing CSI data into a “wireless language” so that LLMs can process and reason about wireless signals in a natural language modeling paradigm, thereby demonstrating early potential for unified modeling across tasks\cite{liu2024llm4cp}.\par
The operational principle of LLMs is based on the Transformer architecture, where large-scale pretraining captures contextual dependencies among tokens \cite{vaswani2017attention}. During training, LLMs learn statistical regularities by predicting the next token, and as model size and data scale grow, emergent abilities such as few-shot reasoning and cross-task generalization arise\cite{ kaplan2020scaling}\cite{ mann2020language}. This pretrain–finetune paradigm allows LLMs to be adapted to new domains, requiring only limited labeled data for task-specific applications.\par
In wireless sensing, LLMs present opportunities in three key aspects. First, by discretizing CSI or RSSI into tokenized “wireless language,” LLMs can provide a unified modeling framework for diverse tasks, avoiding the need for specialized architectures\cite{liu2024llm4cp}. Second, their few-shot capability offers a promising solution for data-scarce sensing tasks such as fall detection, where labeled data are limited. Third, the natural language interface of LLMs provides enhanced interpretability of sensing outcomes, facilitating human–machine interaction and multimodal fusion.\par
Despite these advantages, LLMs face several challenges in wireless sensing. Their effectiveness depends on large-scale training data, yet existing wireless datasets are far smaller and less diverse than natural language corpora, creating a substantial domain gap. Furthermore, LLMs require massive computational resources for training and inference\cite{chowdhery2023palm}, making deployment on resource-constrained sensing devices impractical. Another concern is the hallucination phenomenon, where LLMs generate outputs that are unfaithful to physical reality, posing significant risks for safety-critical sensing applications.\par
To enable broader adoption of LLMs in wireless sensing, several directions have been suggested. Domain-adaptive pretraining can be employed to fine-tune LLMs on CSI or radar data, reducing the domain gap. Hybrid models that combine LLMs with generative approaches such as diffusion models or VAEs may provide both strong generalization and high-fidelity signal generation. Moreover, lessons from multimodal large models such as CLIP\cite{radford2021learning} suggest that jointly modeling wireless, vision, and language data may bring improved interpretability and multi-task adaptability to wireless sensing.

\section{Issues and Challenges}
\label{sec:issue}

\subsection{Data Scarcity}

Wireless sensing systems inherently suffer from the challenge of data scarcity. Unlike vision or speech domains where large-scale datasets are readily available, wireless data collection is often constrained by privacy, environment setup, and hardware heterogeneity. Annotating such data is particularly difficult, as labels typically rely on synchronized sensors or ground-truth systems, which limits offline scalability and requires extensive manual calibration~\cite{xiao2022survey}. 

Although generative AI techniques have been widely proposed to alleviate data scarcity by synthesizing training samples, this strategy faces a fundamental ``chicken-and-egg'' problem: the training of powerful generative models itself demands substantial volumes of high-quality, labeled data. Without sufficient diversity and representativeness in the training set, the generative model may fail to generalize and may even reinforce existing biases in the data distribution~\cite{wen2024generative}.

Recent advances in self-supervised learning (SSL) suggest a promising path forward~\cite{davaslioglu2022self}. By pretraining on vast amounts of unlabeled wireless data, models can learn general representations that transfer effectively to downstream sensing tasks. However, adapting SSL paradigms to the unique nature of wireless signals—such as temporal correlations, multipath propagation, and hardware-induced distortions—remains an open challenge. Unlocking the full potential of generative AI for wireless sensing thus requires a paradigm shift from fully supervised training pipelines toward self-supervised, data-efficient representation learning frameworks~\cite{yang2025revolutionizing}.

\subsection{Model Generalization}

Generalization remains a core bottleneck for wireless sensing systems. Due to the high sensitivity of radio signals to physical environments, device placements, and signal propagation paths, models trained in one environment often perform poorly when deployed elsewhere. While generative AI methods can enrich data diversity through augmentation or simulation, ensuring that the generative models themselves generalize across environments is still a fundamental challenge~\cite{wen2024generative}.

Many existing generative models are environment-specific and struggle to synthesize realistic data when faced with unseen domains. If the generator overfits the source data distribution, the synthesized samples may not meaningfully contribute to cross-domain robustness~\cite{zhou2025dgsense}. Furthermore, the inability to generalize limits the transferability of generative models, which contradicts the goals of scalable deployment.

Some of the existing solutions~\cite{fang2024prism, song2022rf} pretrained with self-supervised objectives across a broad range of environments, offer a potential solution by learning domain-agnostic signal representations~\cite{davaslioglu2022self, alikhani2024large}. However, developing such generalizable generative frameworks requires careful consideration of domain shift, signal variability, and cross-modal consistency. Future research must focus on techniques such as domain adaptation, conditional generation with domain priors, and environment-invariant modeling to bridge this generalization gap.

\subsection{Scalability and Efficiency}

The deployment of generative AI in wireless sensing is hindered by practical concerns regarding model size, inference latency, and computational resource demands. Cutting-edge generative models such as Diffusion and Transformer-based architectures often require significant memory and compute resources, making them ill-suited for real-time applications on edge devices like APs, routers, or mobile terminals.

Although many generative models demonstrate strong performance in controlled offline settings, their scalability to large-scale, real-world deployments remains limited~\cite{zhang2025toward}. Real-time wireless sensing applications, such as activity tracking, anomaly detection, or environment reconstruction, necessitate lightweight, low-latency models capable of operating with constrained power and hardware budgets.

Emerging techniques in parameter-efficient fine-tuning, such as  low-rank adaptation (LoRA)~\cite{hu2021lora, wang2025federated} and adapter-based modules~\cite{llm4wm2025, liu2024few}, offer promising directions for making foundation models practical in wireless scenarios. Additionally, future work should investigate model compression, quantization, and distributed inference to enable scalable generative intelligence at the edge~\cite{dantas2024comprehensive, qu2025mobile}. Balancing generation quality with computational efficiency will be crucial for translating generative AI breakthroughs into deployable wireless sensing solutions.

\section{Future Trend: Wireless Foundation Model}
\label{sec:future}

As the integration of generative AI and wireless sensing continues to deepen, there is an emerging need for a unified, scalable, and adaptable learning paradigm that can generalize across diverse sensing environments and tasks. Inspired by the success of foundation models in NLP~\cite{bommasani2021opportunities}, computer vision~\cite{awais2025foundation}, and speech domains~\cite{ku2025generative}, the concept of a \textit{wireless foundation model} has begun to take shape. Such a model is expected to learn universal representations from large-scale, unlabeled wireless data and adapt efficiently to downstream sensing tasks with minimal supervision and resource cost.

To realize this vision, three enabling technologies are particularly critical: (1) \textit{self-supervised pretraining}, which leverages abundant unlabeled radio frequency (RF) data to learn generalizable features; (2) \textit{efficient fine-tuning}, which enables lightweight adaptation to specific tasks and environments, especially on edge devices; and (3) \textit{cross-modal generation}, which facilitates richer, multimodal environmental understanding by synthesizing or aligning data across sensing modalities. This section elaborates on these three directions and discusses their significance in shaping the next-generation wireless AI systems.

\subsection{Self-supervised Pretraining}
\label{subsec:self-supervised}

Self-supervised learning (SSL) has emerged as a critical tool for training foundational models, especially in domains where labeled data is scarce or difficult to obtain. In wireless sensing, the vast amount of unlabeled RF signal data presents an opportunity for SSL to extract meaningful representations without requiring manual annotation. SSL enables wireless foundation models to learn generalizable features from unlabeled data, facilitating better generalization to unseen environments and tasks. This ability is particularly important for wireless sensing, where the data is often noisy, high-dimensional, and diverse in nature. By learning from this unstructured data, models can uncover underlying patterns that are not easily accessible through traditional supervised learning.

In fields like NLP, CV, and audio processing, SSL has proven to be a game-changer. For example, in NLP, models like BERT \cite{devlin2018bert} and GPT \cite{radford2018improving} leverage SSL to pretrain on vast amounts of text data, enabling them to understand context and semantics in a variety of tasks without requiring explicit supervision. In CV, methods like SimCLR \cite{chen2020simple} and MoCo \cite{he2020momentum} have demonstrated how contrastive learning can help in learning image representations from large-scale datasets without labels. These models have been instrumental in achieving state-of-the-art performance across various downstream tasks. Similarly, in audio processing, models like Wav2Vec \cite{baevski2020wav2vec} have leveraged SSL to learn meaningful audio representations from unlabeled speech data.

In the wireless domain, self-supervised pretraining has started to gain traction. For instance, recent work in \cite{zhang2021unsupervised} explores the SSL techniques for RF signal classification, using unlabeled wireless data to pretrain models before fine-tuning them on specific tasks like gesture recognition or localization. Although these approaches are still in the early stages, they show the potential to create robust models capable of generalizing across different wireless sensing tasks.

To fully unlock the potential of self-supervised pretraining for wireless sensing, future research should focus on improving the scalability of SSL methods to handle the unique challenges of RF data. Additionally, developing more sophisticated SSL techniques that can effectively capture the temporal and spatial dependencies inherent in wireless signals will be critical for advancing the capabilities of wireless foundation models. Furthermore, research into domain-adaptive SSL methods will be essential to ensure these models can transfer knowledge across different wireless environments and applications.

Once fully realized, SSL-based pretraining could dramatically improve the efficiency of wireless AI models, reducing the need for expensive labeled datasets and enabling faster deployment of adaptable and generalizable models in real-world environments.

\subsection{Efficient Fine-tuning}
\label{subsec:efficient-finetuning}

Fine-tuning pre-trained models to adapt to specific tasks has become a standard practice in machine learning. However, in the context of wireless sensing, efficient fine-tuning is essential due to the resource constraints of edge devices, such as APs, routers, or base stations. Wireless foundation models often require adaptation to specific environments (e.g., different RF conditions or device configurations) with minimal computational resources. Techniques like low-rank adaptation (LoRA)~\cite{hu2021lora} and parameter-efficient fine-tuning methods allow for faster, more resource-efficient adjustments to pre-trained models, without the need for retraining the entire model. This is crucial for deploying foundation models in real-time wireless sensing applications, where computational power and energy resources are limited.

In NLP, the introduction of adapter-based fine-tuning methods such as the Adapter-BERT \cite{houlsby2019parameter} has allowed large pre-trained models like BERT to be adapted to new tasks with minimal additional parameters. Similarly, in vision tasks, methods like EfficientNet \cite{tan2019efficientnet} leverage fine-tuning techniques that balance accuracy and computational cost, enabling deployment in resource-constrained environments. LoRA \cite{hu2021lora} has also been applied in NLP and CV to achieve high-quality fine-tuning while reducing the computational overhead by adjusting only a small subset of parameters.

In the wireless sensing field, efficient fine-tuning techniques are still in the exploratory phase. Some studies have started investigating how lightweight adaptation strategies can be employed in wireless models for tasks like localization and activity recognition. For example, a recent work \cite{li2020towards} explored methods for fine-tuning wireless sensing models in a distributed setting, reducing the computational load on edge devices while still maintaining model accuracy.

Future research in wireless sensing should focus on developing more efficient fine-tuning algorithms that are tailored to the specific constraints of edge devices. Incorporating techniques such as sparse fine-tuning, quantization, and knowledge distillation could further optimize model deployment. Additionally, exploring the use of federated learning approaches for fine-tuning wireless foundation models across multiple devices could enable collaborative learning while preserving privacy and reducing computation.

Efficient fine-tuning will allow wireless foundation models to be deployed in a wide range of environments and edge devices, providing high-quality performance while minimizing computational and energy demands.

\subsection{Cross-modal Generation}
\label{subsec:cross-modal}

Cross-modal generation refers to the ability of a model to generate one modality's data from another modality's input, such as generating RF signal patterns from visual data or generating speech from audio signals. In the context of wireless sensing, this capability is transformative as it enables the fusion of multiple sensor types (e.g., RF signals, visual, text, audio) to provide richer and more accurate environmental understanding. Cross-modal generation can enable applications like gesture recognition, environmental mapping, and indoor localization by combining RF data with other sensory inputs like video and sound. This ability to bridge modalities opens up new possibilities for wireless systems, making them more adaptable to dynamic and complex environments.

In CV, the work of \cite{radford2021learning} on Contrastive Language-Image Pre-training (CLIP) demonstrates the power of cross-modal generation, enabling the generation of textual descriptions from images and vice versa. Similarly, in audio processing, models like Speech2Text \cite{wang2020speech2text} have shown the potential of generating speech transcriptions from raw audio signals. In NLP, DALL·E \cite{ramesh2021zero} enables the generation of images from textual descriptions, demonstrating the remarkable potential of cross-modal generation to synthesize complex data across modalities.

% In wireless sensing, research into cross-modal generation is still nascent but shows significant promise. For example, recent studies \cite{li2021wireless} have explored the integration of RF signals and visual data for gesture recognition and activity detection. These studies show that generating complementary information (e.g., 3D position data from RF signals) can help improve the accuracy of perception systems in environments where traditional sensors may fail. However, there are still many challenges in achieving robust and scalable cross-modal generation in wireless systems, particularly in terms of aligning signals across different sensor types and managing the complexity of real-time data generation.

In wireless sensing, cross-modal generation has gained increasing attention, with several recent studies demonstrating its potential to enhance perception in complex environments. For example, \cite{li2021wireless} explored the integration of RF signals and visual data for gesture recognition and activity detection, while \cite{cai2020teaching} proposed a framework to teach RF sensing without using any RF training data by leveraging cross-modal supervision from vision models. Similarly, IMU2Doppler~\cite{bhalla2021imu2doppler} introduced a domain adaptation framework that transfers knowledge from IMU signals to Doppler-based RF sensing models for improved activity recognition. These works show that generating or aligning complementary modalities—such as visual, inertial, or spatial priors—can significantly improve system robustness, especially in conditions where individual sensors may fail. Nonetheless, achieving robust and scalable cross-modal generation remains challenging, particularly in terms of synchronizing heterogeneous sensor signals and handling the complexity of real-time generative inference in dynamic scenarios.

To realize the full potential of cross-modal generation in wireless sensing, future research should focus on improving the alignment between different sensor modalities and developing efficient generative models that can handle the diverse and noisy nature of wireless data. Additionally, exploring new generative architectures, such as GANs and diffusion models, could improve the quality and reliability of generated signals. Another area of focus will be integrating generative models into real-time wireless sensing systems, ensuring that models can work efficiently in dynamic, real-world environments.

Cross-modal generation holds immense potential to revolutionize wireless sensing applications, making systems more robust, adaptive, and capable of handling complex, multi-sensor environments. As these models evolve, they will enable new applications such as more accurate environmental mapping, localization, and gesture recognition, bridging the gap between different types of wireless sensors.

\section{Conclusion}
\label{sec:conclusion}

Generative Artificial Intelligence (GenAI) has demonstrated great potential in enhancing wireless sensing systems. In this paper, we explored the integration of GenAI into wireless sensing pipelines, focusing on two primary integration modes: as a plugin to augment traditional models, and as a standalone solver to directly address sensing tasks. We examined a range of generative models, discussing their strengths and limitations in the context of various wireless sensing applications. Additionally, we analyzed the key issues and challenges in integrating GenAI with wireless sensing, identifying them as critical open problems within the wireless sensing community. Building on these challenges, we propose the future trend of building wireless foundation model to support scalable, adaptable, and efficient wireless sensing systems.


% \section{References Section}
% You can use a bibliography generated by BibTeX as a .bbl file.
%  BibTeX documentation can be easily obtained at:
%  http://mirror.ctan.org/biblio/bibtex/contrib/doc/
%  The IEEEtran BibTeX style support page is:
%  http://www.michaelshell.org/tex/ieeetran/bibtex/
 

\bibliographystyle{IEEEtran}
\bibliography{bib/reference}

\vfill

\end{document}